\documentclass[10pt,journal,compsoc]{IEEEtran}

\ifCLASSOPTIONcompsoc
  \usepackage[nocompress]{cite}
\else
  \usepackage{cite}
\fi

\ifCLASSINFOpdf
\else
\fi
\usepackage{subfigure}
\usepackage[lined,ruled,noend]{algorithm2e}
\usepackage[caption = false]{subfig}
\usepackage{tabularx}
\usepackage{multicol}
\usepackage{multirow}
\usepackage{booktabs}
\usepackage{graphicx}
\usepackage{xcolor}
\usepackage{diagbox}
\usepackage{threeparttable}
\usepackage{amsmath}
\usepackage{multirow}
\usepackage[normalem]{ulem}
\usepackage[version=4]{mhchem}

\usepackage{mathrsfs}
\usepackage[mathscr]{euscript}
\usepackage{circledsteps}
\usepackage{ulem}
\usepackage{ragged2e}
\usepackage{float}
\usepackage{tabularx}

\usepackage{amssymb}

\begin{document}
 
\title{Reconstructing Randomly Masked Spectra Helps DNNs Identify Discriminant Wavenumbers}
\author{Yingying~Wu,
        Jinchao~Liu,
        Yan~Wang,
        Stuart~Gibson,
        Margarita~Osadchy 
        and~Yongchun~Fang, \textit{Senior member}, IEEE
\IEEEcompsocitemizethanks{\IEEEcompsocthanksitem Y. Wu, J. Liu and Y. Fang are with the Institute of Robotics and Automatic Information System (IRAIS), College of Artificial Intelligence, Nankai University, China. e-mail: wuyy@mail.nankai.edu.cn, \{liujinchao, fangyc\}@nankai.edu.cn (Corresponding author: J. Liu) 
\IEEEcompsocthanksitem Y. Wang is with VisionMetric Ltd, Canterbury, Kent, UK. e-mail: yanwangnott@gmail.com
\IEEEcompsocthanksitem S. Gibson is with the School of Physics and Astronomy, University of Kent, UK. e-mail: s.j.gibson@kent.ac.uk
\IEEEcompsocthanksitem M. Osadchy is with the Department of Computer Science, Haifa University, Israel. e-mail: rita@cs.haifa.ac.il

\IEEEcompsocthanksitem  \copyright 2024 IEEE. Personal use of this material is permitted.  Permission from IEEE must be obtained for all other uses, in any current or future media, including reprinting/republishing this material for advertising or promotional purposes, creating new collective works, for resale or redistribution to servers or lists, or reuse of any copyrighted component of this work in other works.

\IEEEcompsocthanksitem This work has been published in the IEEE Transactions on Pattern Analysis and Machine Intelligence, vol. 46, no. 5, pp. 3845-3861, May 2024, doi: 10.1109/TPAMI.2023.3347617.
}
}

\markboth{IEEE TRANSACTIONS ON PATTERN ANALYSIS AND MACHINE INTELLIGENCE}%
{Shell \MakeLowercase{\textit{et al.}}: Bare Demo of IEEEtran.cls for IEEE Journals}

\IEEEtitleabstractindextext{%
\begin{abstract}
\justifying{
Nondestructive detection methods, based on vibrational spectroscopy, are vitally important in a wide range of applications including industrial chemistry, pharmacy and national defense. 
Recently, deep learning has been introduced into vibrational spectroscopy showing great potential. Different from images, text, etc. that offer large labeled data sets, vibrational spectroscopic data is very limited, which requires novel concepts beyond transfer and meta learning. To tackle this, we propose a task-enhanced augmentation network (TeaNet). The key component of TeaNet is a reconstruction module that inputs randomly masked spectra and outputs reconstructed samples that are similar to the original ones, but include additional \textcolor{black}{variations} learned from the domain. These augmented samples are used to train the classification model. The reconstruction and prediction parts are trained simultaneously, end-to-end with back-propagation. Results on both synthetic and real-world datasets verified the superiority of the proposed method. In the most difficult synthetic scenarios TeaNet outperformed CNN by 17\%. We visualized and analysed the neuron responses of TeaNet and CNN, and found that TeaNet’s ability to identify discriminant wavenumbers was excellent compared to CNN. Our approach is general and can be easily \textcolor{black}{adapted} to other domains, offering a solution to more accurate and interpretable few-shot learning.
}

\end{abstract}

\begin{IEEEkeywords}
Masked CNN, deep learning, vibrational spectroscopy
\end{IEEEkeywords}}

\maketitle
\IEEEdisplaynontitleabstractindextext
\IEEEpeerreviewmaketitle

\IEEEraisesectionheading{\section{Introduction}\label{sec:introduction}}
\IEEEPARstart{V}{ibrational} spectroscopy, including infrared and Raman, has been widely used for identification and quantification of solids (particles, pellets, powers, films, fibers), liquids (gels, pastes) and gases.  Infrared spectroscopy makes use of the vibrational transitions of a molecule, provides fast, non-contact, and non-destructive analysis and has found a wide range of applications in a variety of industries such as petrochemical, chemical, pharmaceutical, cosmetic, food.

\begin{figure}
	\centering
    \subfigure{
	\includegraphics[width=0.5\textwidth]{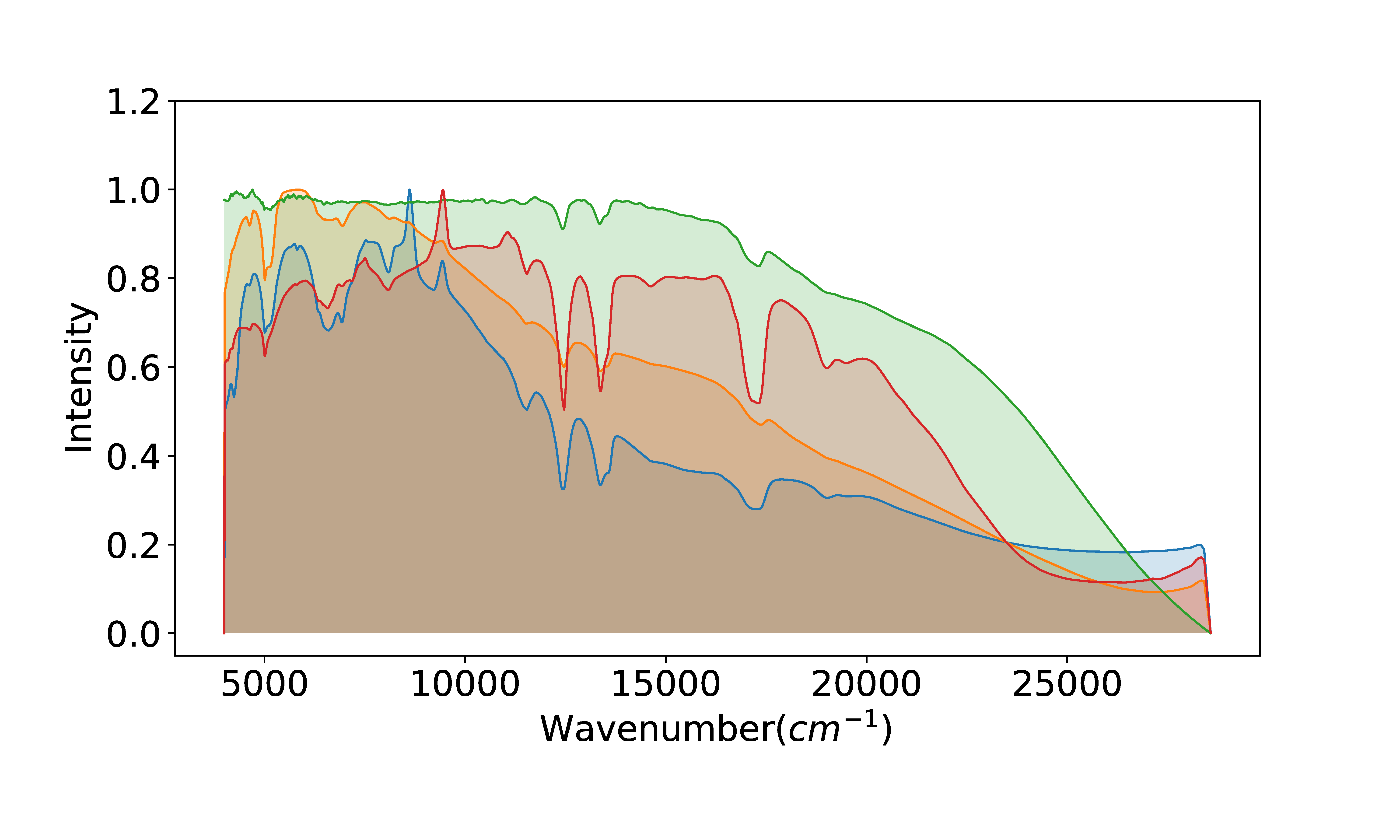}
    }\vspace{-0.5cm}
    \subfigure{
    \includegraphics[width=0.5\textwidth]{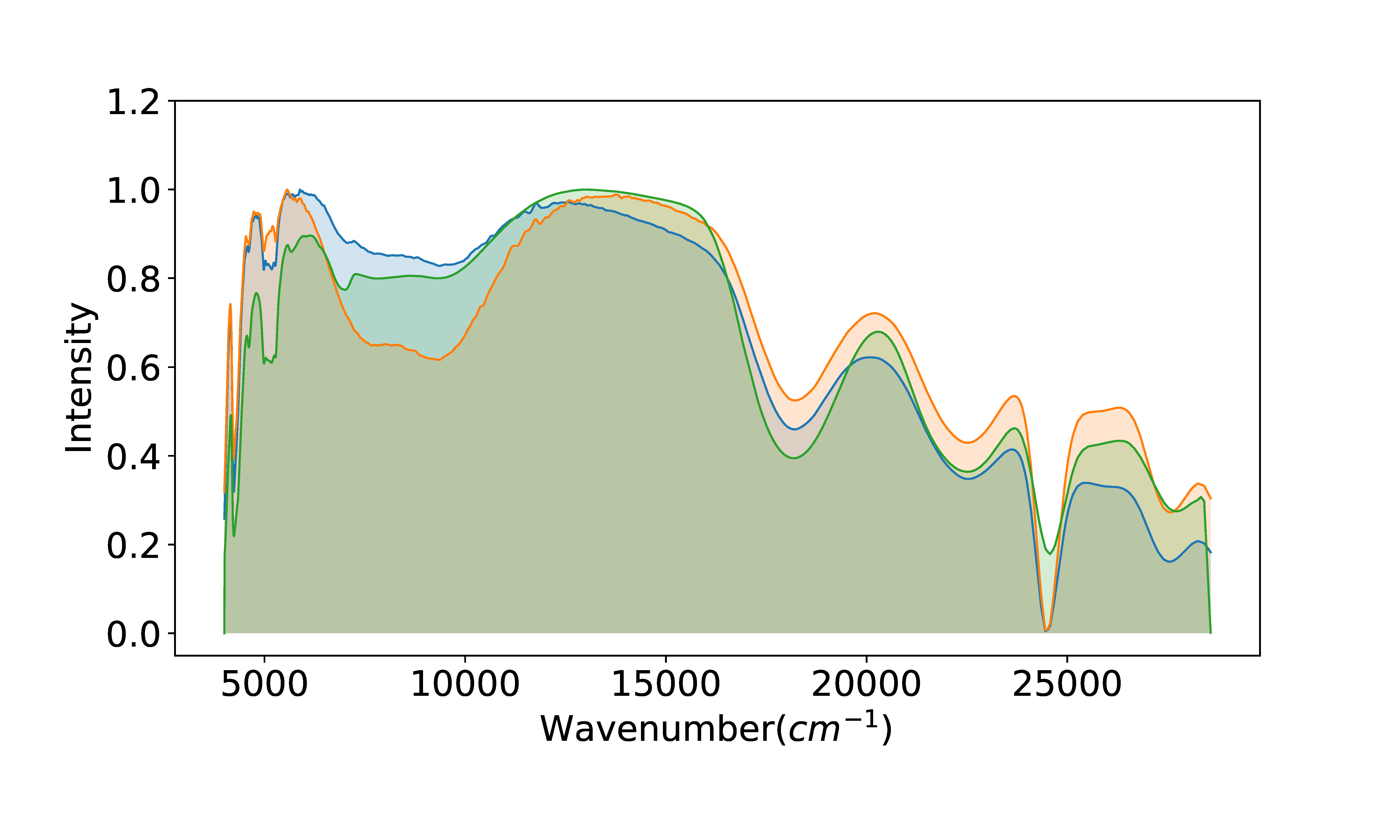}
    }\vspace{-0.5cm}
	\caption{Examples of near-infrared spectra of two minerals, \textit{Fluorapatite
} and \textit{Rhodochrosite}, from the USGS dataset. For each mineral, there are only a handful of samples which differ dramatically from each other. Recognising these minerals from their spectra falls into the category of ``few-shot learning'' which is rather challenging due to the lack of training samples and the complicated nature of infrared spectra.}
	\label{Fig:nir_examples}
\end{figure}

An infrared spectrum captures unique molecular information that can be used to characterize a substance and is therefore often referred to as ``molecular fingerprint''. Examples of infrared spectra of minerals are shown in Fig.~\ref{Fig:nir_examples}. It can be seen that decryption of such spectra is non-trivial, especially in the presence of impurity in the sample and noise caused by the acquisition process. Typically, the analysis of spectra relies heavily upon machine learning methods \cite{bouveresse1996improvement,liu2017deep,LIU2019Dynamic,Cui2018,Cui2019}.

\begin{figure*}
	\centering
    \subfigure[A Spectrum]{
    \includegraphics[width=0.275\textwidth]{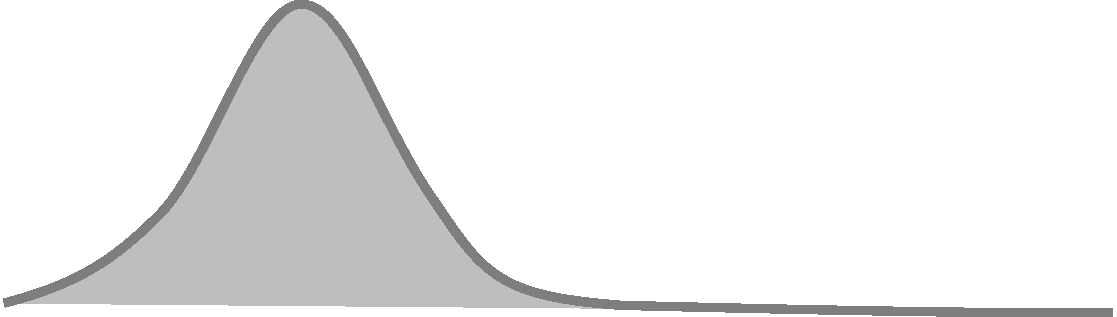}
    }
    \subfigure[Randomly Masking]{
    \includegraphics[width=0.275\textwidth]{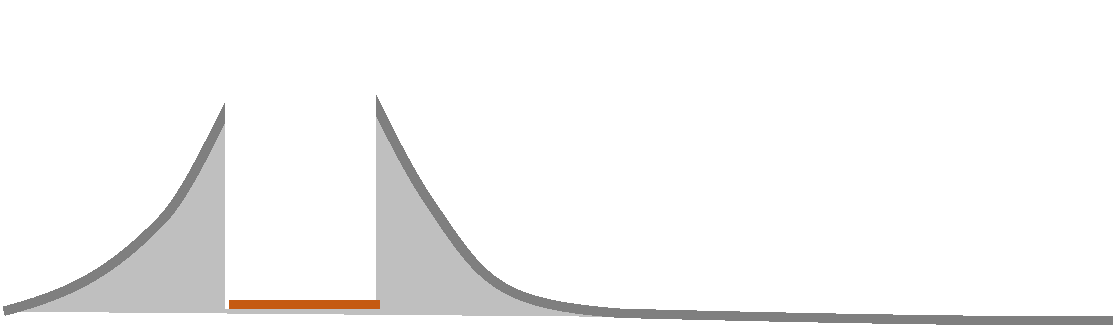}
    }
    \subfigure[CutMix~\protect\cite{yun2019cutmix}]{
    \includegraphics[width=0.275\textwidth]{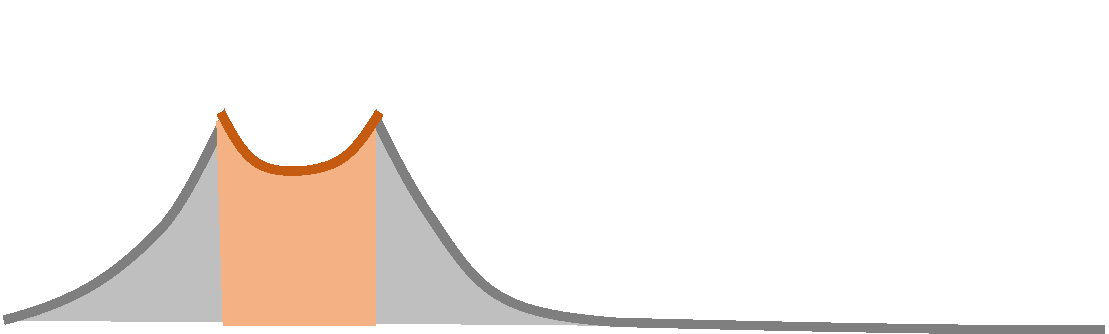}
    }
    \subfigure[MixUp~\protect\cite{zhang2017mixup}]{
    \includegraphics[width=0.275\textwidth]{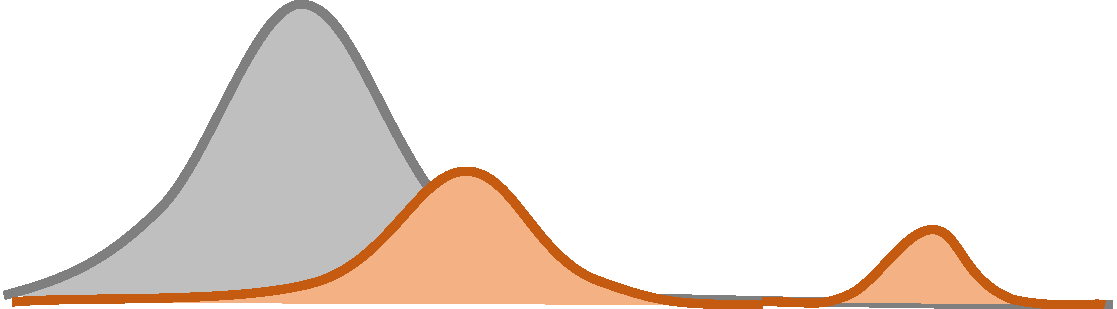}
    }
    \subfigure[DAGAN~\protect\cite{antoniou2017data}]{
    \includegraphics[width=0.275\textwidth]{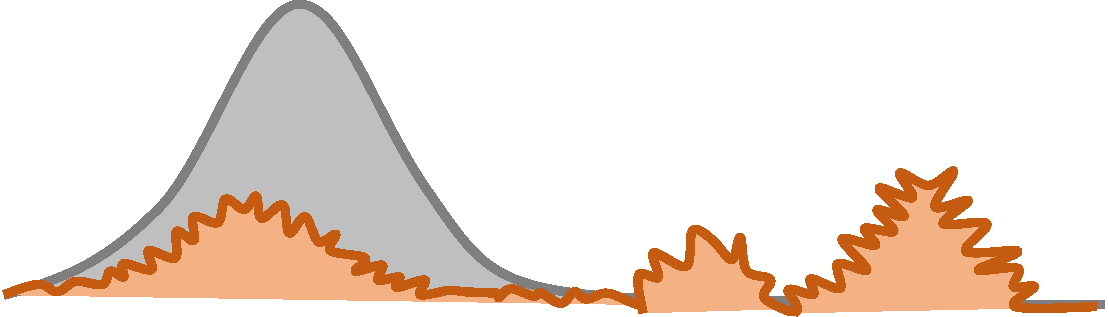}
    }
    \subfigure[TeaNet (ours)]{
    \includegraphics[width=0.275\textwidth]{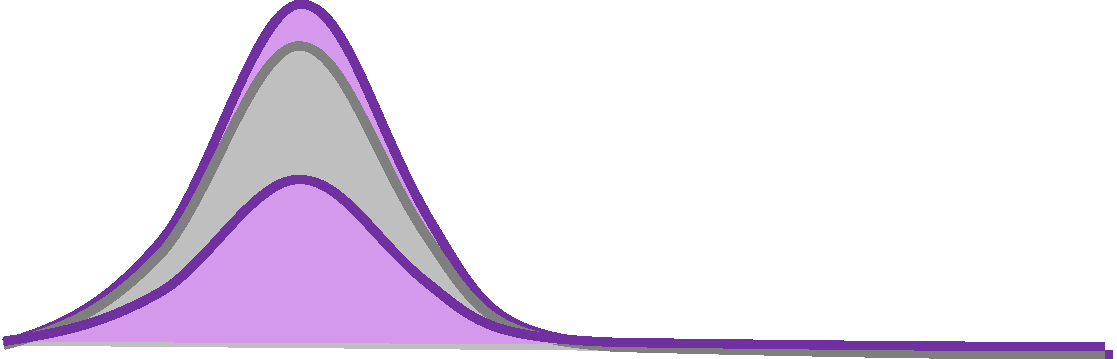}
    }
	\caption{
	\textcolor{black}{
	Illustration of typical augmentation behaviors of existing popular data augmentation strategies and the proposed TeaNet. The variations generated by the existing methods and the proposed TeaNet are indicated in orange and purple respectively. (a) A spectrum with one characteristic peak. (b) Randomly masking generates spectra where a random segment is dropped. (c) CutMix masks the spectrum which is then filled with a random segment from other sample. (d) MixUp combines multiple spectra linearly and conceptually may be regarded as a merging of masking and filling in one step. (e) GAN trained with small data tends to generate augmented sample with no or false variations due to the insufficient training (as observed in our experiments). (f) Our approach TeaNet is able to generate artificial samples with sensible variations.}}
	\label{Fig_illustration_all_methods}
\end{figure*}

\textcolor{black}{Traditionally, automatic recognition of infrared spectra has been based on conventional machine learning methods such as kNN, SVM, PLS} \cite{KNN_Analyst,Sattlecker2010,PLSDA_review2018}. These methods often \textcolor{black}{require pre-processing that typically includes some or all of the following steps: smoothing, baseline correction, multiplicative scatter correction (MSC), standard normal variate (SNV) correction, wavenumber selection and principal components analysis (PCA)} ~\cite{Zhang2010,Peng2011,HelinPreprocessing2021}. Such pre-processing in most cases is non-trivial and could erroneously remove useful features while retaining undesirable noise in the signal.

In recent years, deep neural networks have been \textcolor{black}{proposed for vibrational spectroscopy that have significantly outperformed the conventional machine learning methods} in many applications including mineral classification~\cite{liu2017deep,Liu2018}, chemical analysis~\cite{Liu2018,Fan2019}, pathogenic bacteria identification~\cite{Ho2019}, rapid detection of COVID-19 causative virus (SARS-CoV-2)~\cite{RamanForCovid2021,RamanForCovid_Water_2021,RamanForCovid_Water_2022}, metabolite gradients monitoring~\cite{Lussier2019}, cytopathology~\cite{Kraub2018}, soil properties prediction~\cite{Liu2018,Padarian2019}, mine water inrush~\cite{Hu2019}, etc. \textcolor{black}{The use of deep networks in analysing spectra has not only improved the accuracy, but also removed the need for manual, non-trivial, and time consuming preprocessing of spectra.}

In many real-world applications of vibrational spectroscopy, there are only a handful of samples/spectra available for each substance. As training deep networks usually requires very large datasets, applications with small data regimes are especially difficult for deep networks. To tackle this problem, previous work proposed and adopted data augmentation techniques~\cite{liu2017deep,Bjerrum2017}, employed transfer learning~\cite{Liu2018,SeddikiNatureComm} or meta-learning~\cite{LIU2019Dynamic}. Despite these efforts, the success of large neural networks in few-shot spectrum recognition has been limited.  

\textcolor{black}{Generative adversarial networks (GANs) have shown to be a powerful tool for learning distributions from examples and generating artificial samples. GANs have been used for data augmentation~\cite{antoniou2017data,luo2018eeg,kong2020physgan} in various tasks.} However, we show here that state-of-the-art GANs fail in improving few-shot spectrum recognition (see Table \ref{Tab:Main_Benchmark_Results}). \textcolor{black}{This is expected as training GANs requires even more data than training classifiers.} This result motivated us to develop an alternative \textcolor{black}{generative} model that trains with self-supervision and makes \textcolor{black}{ better use of the small training data (i.e. small volumes of training data).} 

Besides GANs, there exist a number of popular data augmentation strategies involving masking a signal such as Random Erasing~\cite{zhong2020random} and CutMix~\cite{yun2019cutmix}. These methods mask random regions of samples which are then \textcolor{black}{``roughly''} filled with zeros or random patches from other samples. It has been shown that they are quite effective when a sufficient number of training samples are available. However, as shown in our experiments, when only a handful of samples are available, these strategies are of little use. When large data is available, introducing a relatively small amount of roughly-filled masked samples could help regularizing the model. In the case of small data where true variations are limited, it is essential to generate artificial samples with sensible variations, instead of near-arbitrary mixups. Fig.~\ref{Fig_illustration_all_methods} shows a graphical illustration of typical variations generated by existing data augmentation strategies and the proposed TeaNet.

\textcolor{black}{Inspired by these augmentation strategies including Random Erasing and CutMix which can be viewed as (random) masking followed by a rough operation of filling, we} take this one step further and propose to use a generative model to reconstruct the randomly masked regions to create artificial samples with sensible variations, instead of rough fillings. With the task of \textcolor{black}{``mask-reconstruct''}, we can create from a single spectrum many training sample pairs to train the generative model. This is especially useful in small data applications. We named our novel method \textit{Task-Enhanced Augmentation Network} (TeaNet). The  mask-reconstruct approach was also inspired by masked language model training  BERT~\cite{Devlin2019BERTPO} and the inpainting task in computer vision~\cite{InpaintingReview2020, Pathak2016, UNet-liu2018image}.

\subsection{Contributions}
We propose a new method for few-shot classification \textcolor{black}{based on a novel} tasked-enhanced augmentation network. Our approach outperforms previous state-of-the-art methods in infrared spectra classification and it offers interpretability by selecting the most discriminant parts of the \textcolor{black}{input (defined over a range of wavenumbers)}. Specifically, 

\begin{enumerate}
    \item We propose TeaNet, a task-enhanced augmentation network that learns to produce novel samples by filling in the masked part of the input signal \textcolor{black}{with the goal of maximizing the accuracy of the classifier. }\textcolor{black}{Source codes of TeaNet are available at \underline{https://github.com/Chaoscendence/TeaNet}}.
    \item Our experiments on synthetic and real-world infrared spectrum datasets showed that TeaNet significantly outperforms the state-of-the-art methods in the field. 
    \item We analysed TeaNet and other methods, and showed that TeaNet possessed an \textcolor{black}{excellent ability for discriminating between spectra that were very similar to one another. Our visualization showed that TeaNet was able to locate discriminant regions with much better accuracy than CNN, and therefore offers much better interpretability in addition to superior classification performance.}
\end{enumerate}
\textcolor{black}{The main idea behind the TeaNet is general and the architecture can be easily adapted to other domains, offering a general approach to more accurate and interpretable few-shot learning.}
\subsection{Structure of the Paper}The remainder  of  this  paper  is  organized  as  follows:  Section \ref{sec:related_work} reviews the related  work.  Section \ref{sec:TeaNet} describes  the proposed task-enhanced augmentation network (TeaNet). Section \ref{sec_experiment_onsyntheticspectra} and \ref{sec:experiments_realdata} present and discuss the experiments on synthetic data and real-world benchmark datasets. Section \ref{sec:conclusion} summarises our work and discusses potential directions for future research.

\section{Related work}\label{sec:related_work}
Our work addresses the few-shot learning by augmentation and proposes a generation network which creates synthetic samples that are reconstructed from the real masked samples. It is trained by minimizing a combination of reconstruction and classification losses. We discuss the relevant work in the related areas, specifically, self-supervision through masking and augmentation.

\begin{figure*}[!htb]
\centering
\includegraphics[width=\linewidth]{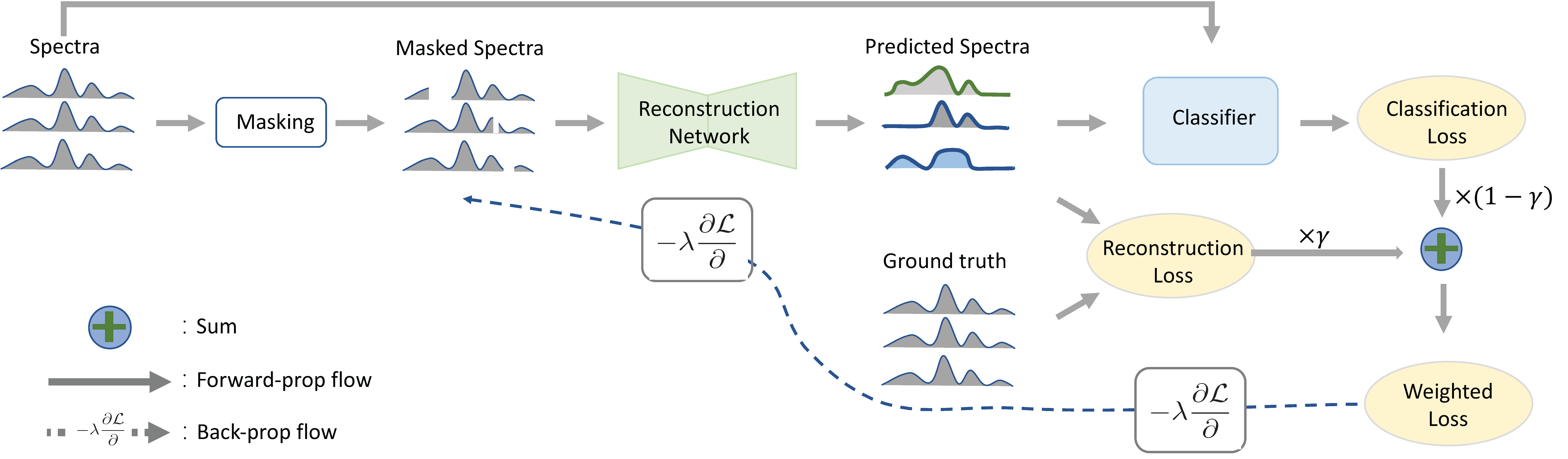}
\caption{Diagram of the proposed method TeaNet. It consists of two neural networks, namely reconstruction network and a differential classifier for a given application. The reconstruction network is responsible for generating augmented samples by repairing the masked/corrupt spectra and being led by the classification results. The trainable weights in the reconstruction network are optimised against the weighted sum of the classification and reconstruction losses. While the classifier is trained towards minimizing the classification loss only.}
\label{fig:TeaNet_diagram}
\end{figure*}

\begin{figure}[!htb]
	\centering
	\includegraphics[width=0.475\textwidth]{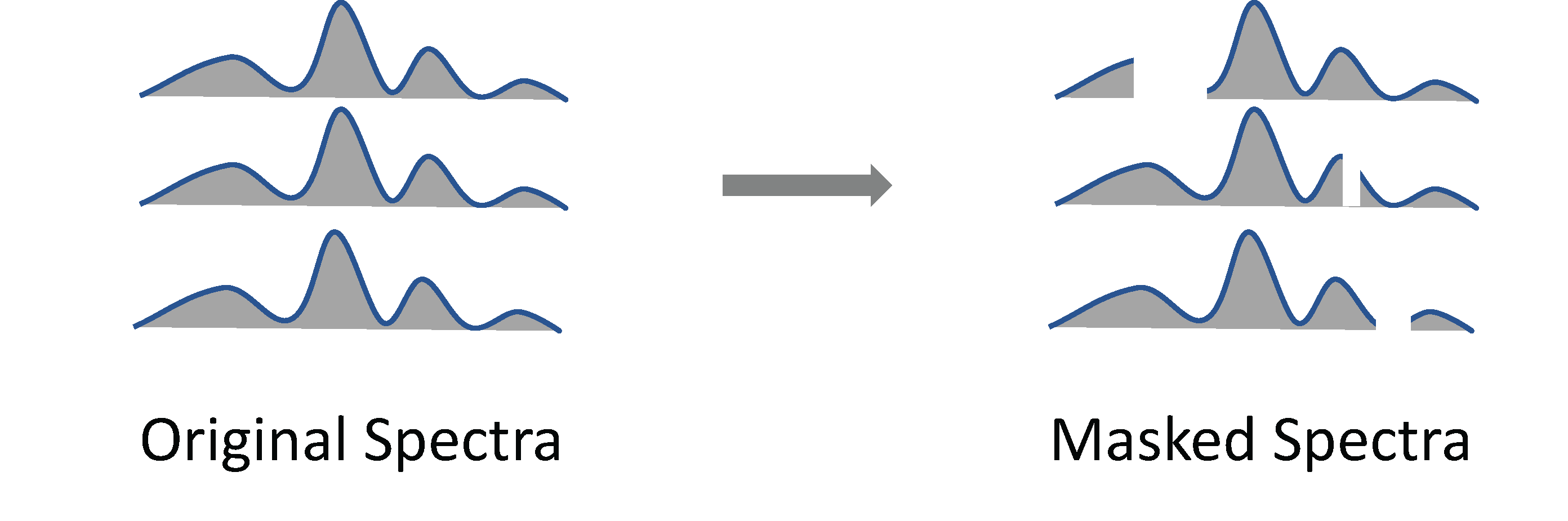}
	\caption{Examples of masked spectra and the corresponding original ones. Similar to generating pairs to train Siamese networks for few-shot learning\cite{LIU2019Dynamic}, generating lots of original-masked spectrum pairs allow us to exploit the training samples in a novel way and thus may help train larger model and achieve better accuracy in the case of few-shot spectrum recognition.}
	\label{Fig:masked_examples}
\end{figure}

\subsection{Data Augmentation}
A straightforward solution to the problem of insufficient training data is to synthesize new samples by applying various transforms to the real samples. Simple geometric transforms such as flipping, scaling, rotation and shifting are often applied to image data, whilst patch sampling \cite{krizhevsky2017imagenet,kang2017patchshuffle}, sample mixing \cite{inoue2018data,summers2019improved} and random erasing and occlusion \cite{zhong2020random,devries2017improved} are among other common strategies. 
However, all these augmentation methods only involve low-level image operations and do not account for semantic knowledge of the content of the data. 
We show in our experiments (Section~\ref{sec:experiments_realdata}) that augmenting training of a deep classification network with spectra obtained by low-level signal operations (we refer to it as CNN-Partial in our experiments) was not always beneficial and even worsened the results in some cases. 

{\color{black}
Instead of applying low-level transformations,} Wang et al.\cite{wang2021regularizing} proposed synthesizing new samples at semantic levels by manipulating features. Since features at higher layers of a deep network are characterized by better abstraction, one can assume that introducing some variation at the feature level could offer augmentation with semantic variation.  

In recent years, GAN models have shown \textcolor{black}{ great capability in synthesizing image data (especially facial images \cite{karras2017progressive}) with high fidelity.} GAN models with special architectures \textcolor{black}{have also} demonstrated potential for data augmentation \cite{antoniou2017data,zhu2017unpaired,luo2018eeg}. 
The CWGAN \cite{luo2018eeg} was proposed to augment the electroencephalogram (EEG) data for emotion recognition. It combines WGAN and CGAN and feeds a class label as an additional input to both the generator and discriminator to synthesize samples of specific classes. Antoniou et al. \cite{antoniou2017data} developed DAGAN which consists of a CGAN generator and an improved WGAN critic network. The generator takes a real data item and a random vector as inputs and synthesizes a new data item through an encoder-decoder pipeline. The random vector is sampled from a multi-variate Gaussian distribution to encode the intra-class variations. The generated and the input samples form a combined fake item, while the input sample and another real sample from the same class form a combined real item. The discriminator is trained to distinguish the fake from the real. After training, the original data is then augmented to train a classifier for the target task. 

Data augmentation in the feature domain was considered to assist a few-shot learning. \textcolor{black}{Different} from our application,~\cite{GaoSZZC18} makes an assumption on the availability of a large set of labeled data that is related/similar to classes with the small training data.  A special case of GAN is trained  to model the latent distribution of each novel class given examples from  similar ``large'' classes. 
The work in~\cite{SchwartzKSHMKFG18} uses an auto-encoder to learn differences between same-class pairs of training examples in the latent space, and then applies these ``deltas'' to small novel data to synthesize samples from the new class. The work in~\cite{WangGHH18} also performs augmentation in the latent space, but the augmentation is guided by the classification loss, which is  similar to our method. However, our overall approach is very different from~\cite{WangGHH18}: in our method  we create new samples (not features) by reconstructing the masked regions using a reconstruction loss in addition to the classification loss.

\subsection{Self-supervised learning}
Self-supervision via reconstruction of partially masked inputs has been widely used in language modeling e.g., BERT~\cite{Devlin2019BERTPO} and more recently in computer vision e.g., ViT~\cite{abs-2010-11929}. The main goal of these architectures is to utilize the large unlabeled data for pre-training the encoder and then fine-tuning it for the downstream tasks. 

A recent work in~\cite{abs-2111-06377}  proposed an asymmetric architecture for masked auto-encoder in which the encoder inputs a small number of randomly selected image patches from an input image with positional information, while the decoder receives the embeddings and the masked symbols and reconstructs the missing parts. It is suggested that masking a very large number of  patches in the input encourages the encoder to learn semantic information and not just hole filling. The specific architecture makes the training more efficient.  While our approach also employs masking for self-supervision of the reconstruction network, it is used in combination with the classification loss. We do not use the encoder after training, the reconstruction network in our approach is used to enlarge small training data by adding variation to the existing samples via generating the masked parts. 

LTSA \cite{jenni2018self} is a self-supervised feature learning framework where a damage-and-repair network learns to synthesize images with mild artifacts by randomly masking the latent representation of input images, and a discriminator is trained simultaneously to differentiate the images with artifacts from normal ones. The features extracted by the discriminator can later be used for downstream classification tasks. \textcolor{black}{Note that the damage and repair network is relevant to, but different from, our approach.} In LTSA, masking or damaging was applied to the latent representation, instead of the input images, in order to generate global artifacts. The success of this scheme greatly relies on good representation to be learned before it can take effect. Our experiments showed that in the case of few-shot learning this approach performed significantly worse than the baseline methods, as shown in Table \ref{Tab:Main_Benchmark_Results}.

\section{TeaNet: Task-enhanced Augmentation Networks}\label{sec:TeaNet}
The small training data problem has been traditionally approached by transfer learning and by meta-learning~\cite{Fewshot_Koch2015SiameseNN,Fewshot_MatchingNet,Fewshot_PrototypeNet,Fewshot_FinnAL17}. We cannot use either of these approaches in the infrared spectrum domain, as transfer learning assumes the existence of a large labeled training set in a similar domain, which  does not \textcolor{black}{exist} in our application; meta-learning assumes a  data set with many episodes from a similar domain and is limited to a relatively small classification problems (with 5-50 ways). In infrared spectrum recognition the classification problem is usually large, including hundreds of classes. 

To tackle the problem of few-shot \textcolor{black}{learning with a large number} of classes and no auxiliary training data, we choose data augmentation using generative models. This is a non-trivial task, as training a good generative model with no prior information requires large training sets that we do not have in our setting.  This makes regular GANs fail as experimentally confirmed in Table \ref{Tab:Main_Benchmark_Results}. One way to approach this problem is to impose additional constraints on the generative model to reduce its need for a large number of training samples.  

Inspired by masking ideas in  pre-training large transformers in language modeling~\cite{Devlin2019BERTPO}, and computer vision~\cite{abs-2010-11929,abs-2111-06377} as well as the inpainting task in images~\cite{InpaintingReview2020, Pathak2016, UNet-liu2018image}, we propose task-enhanced augmentation networks (TeaNet), where a generative model learns through reconstructing samples from randomly masked ones. Note that the masking approach in previous work was used to pretrain the encoder on a very large data and then fine-tune it to the downstream tasks. In this work, we take a different approach and train an auto-encoder not only to reconstruct masked inputs, but to fill the gaps, such that the repaired sample minimizes the classification loss. We augment the original sample with a number of masked samples by randomly masking the original one, as shown in Fig.~\ref{Fig:masked_examples} and repairing them using the auto-encoder. 
The repaired samples are passed to the classifier and the classification loss on these \textcolor{black}{samples is used to} train simultaneously  the classifier and the auto-encoder (in addition \textcolor{black}{to the reconstruction loss)}. A graphical illustration can be found in Fig.~\ref{fig:TeaNet_diagram}.
  
The success of masking in training language models was attributed to the fact that languages are human-generated, and thus
are highly semantic and information-dense~\cite{abs-2111-06377}. Masking words is enough to facilitate the model to capture the semantics of the language.  In images, due to spatial redundancy, a missing patch can be recovered from neighboring patches without capturing much of the semantics, thus requiring very heavy masking~\cite{abs-2111-06377}. The question is whether the masking approach would work in spectroscopy and what should be the masking strategy here?
The spectroscopy theory in physics shows that any part of an infrared spectrum is related to other parts, which indicates that inferring a missing/masked part from the rest of an infrared spectrum is indeed physically possible, though is very challenging \cite{BookSpectroscopyShen}.

\begin{figure}[!tb]
\centering
\subfigure[]{
\includegraphics[width=0.475\columnwidth]{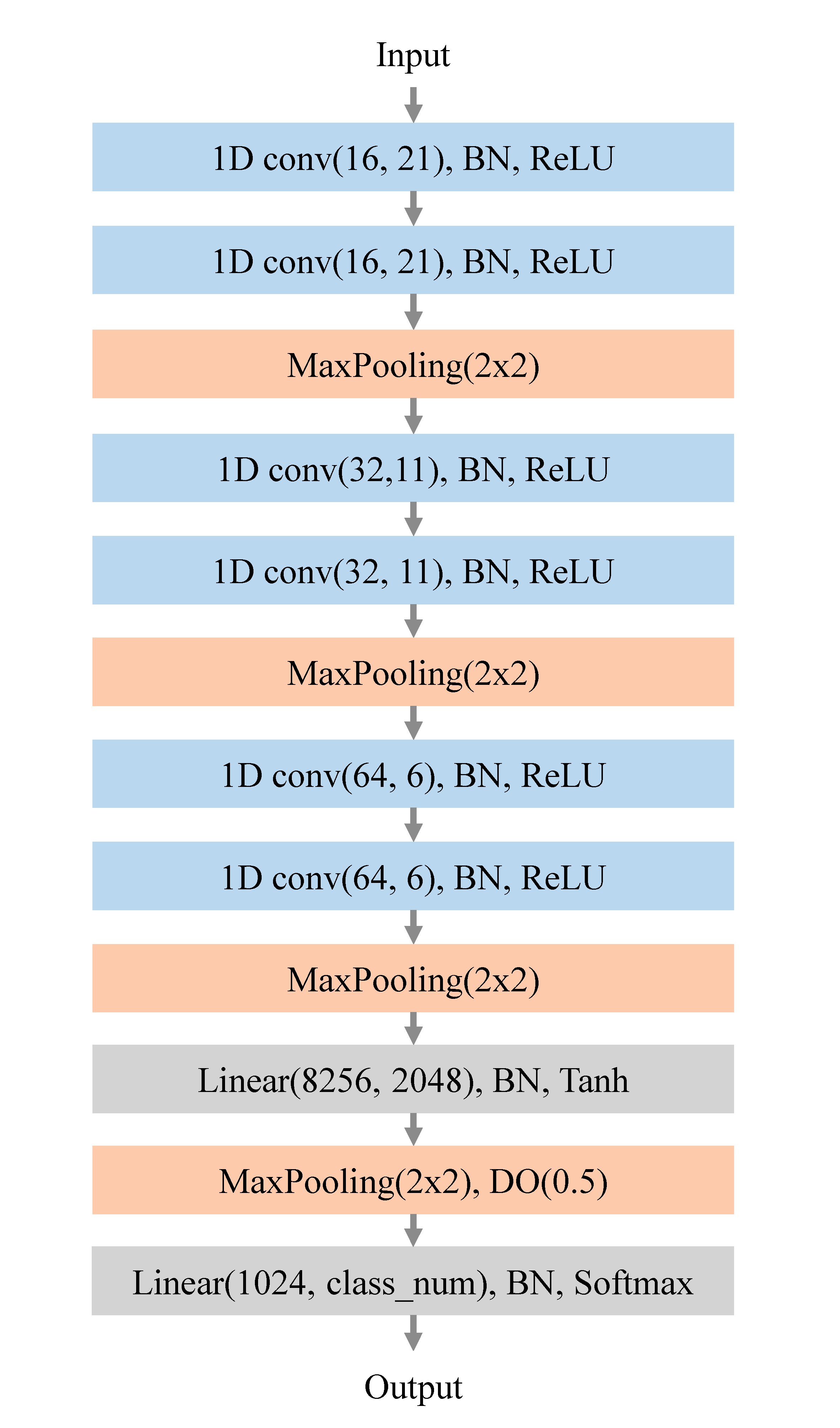}
}
\subfigure[]{
\includegraphics[width=0.4\columnwidth]{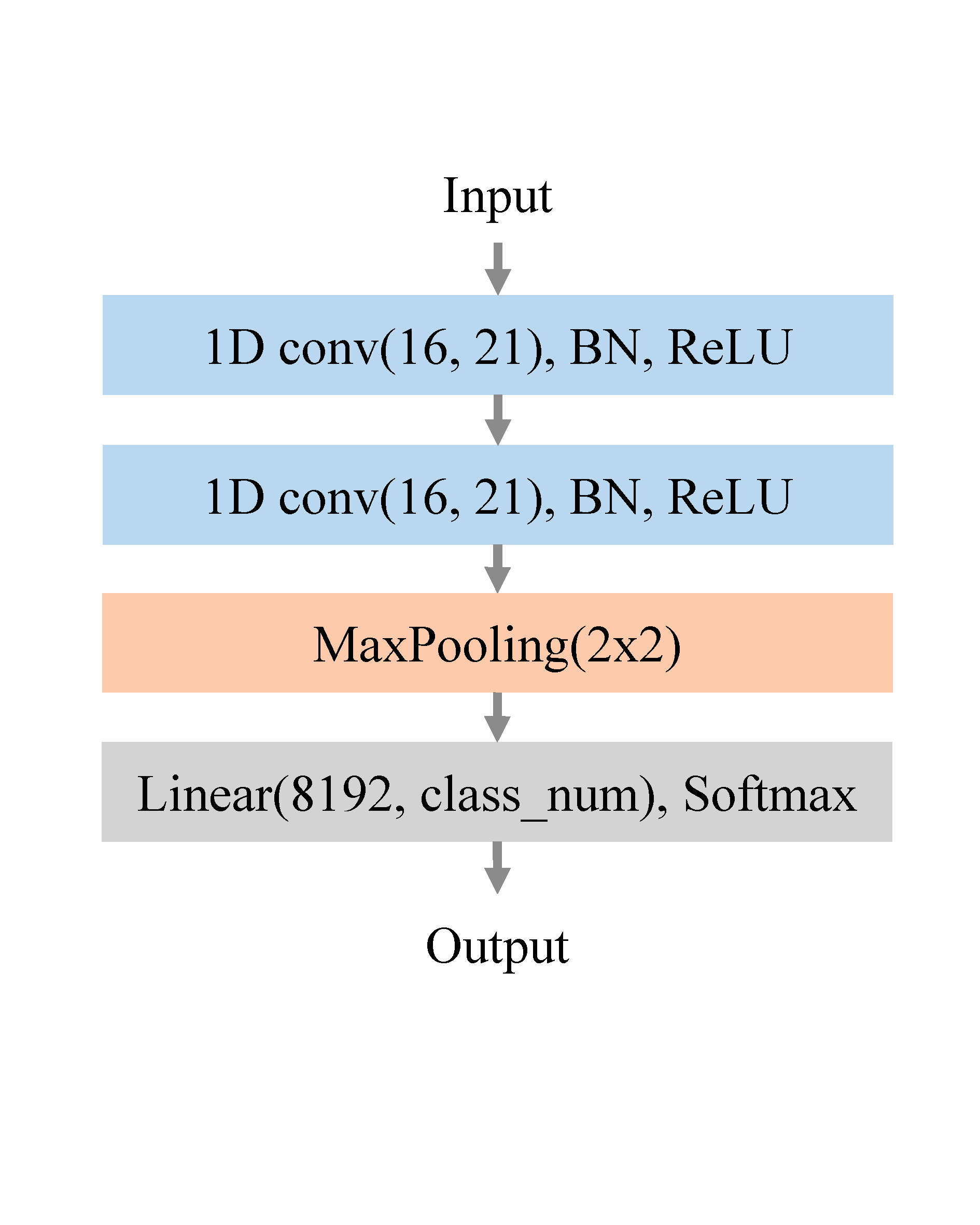}
}
\caption{\textcolor{black}{Network architectures of the classification modules for TeaNet used in the real-world and synthetic datasets. BN stands for batch normalization and DO stands for Dropout.} }
\label{fig:classification_details}
\end{figure}

\subsection{Network Architecture}\label{Sec:Subsection_network_architecture}
Our network consists of two modules, namely the \textit{reconstruction} module and the \textit{classification} module, which are combined in a pipeline. The classification module is responsible for a spectrum recognition task. Any differentiable model, e.g., convolutional networks, can be used for the classification module. 
We employ double-pyramid LeNets that \textcolor{black}{have been shown to be} very effective in spectrum recognition~\cite{liu2017deep,LIU2019Dynamic} in our experiments. The detailed architectures especially the number of layers were optimised for different applications (datasets) as shown in Fig.~\ref{fig:classification_details}.

\begin{figure}[!tb]
\centering
\includegraphics[width=0.9\columnwidth]{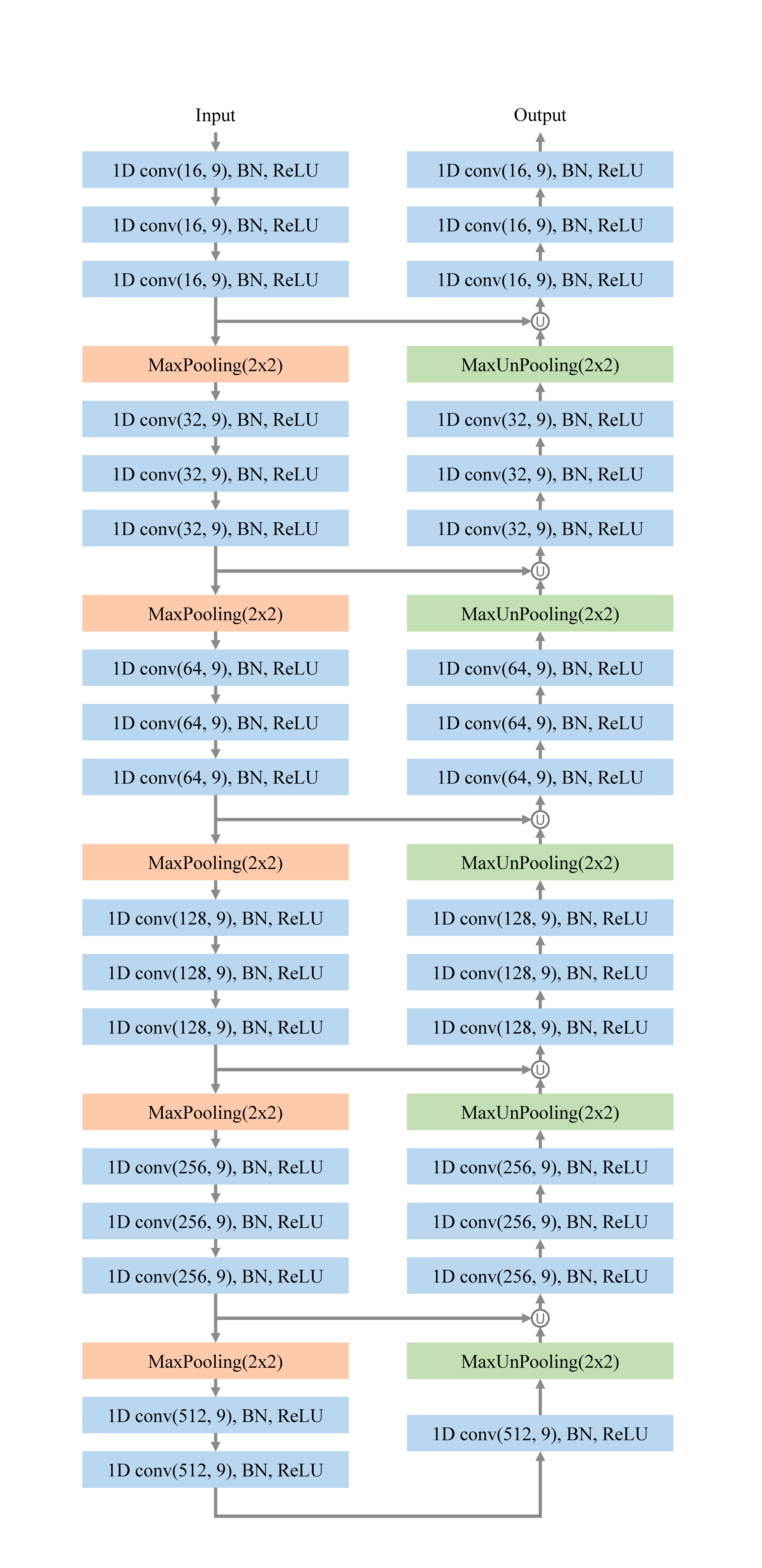}
\caption{\textcolor{black}{Network architecture of the reconstruction module in the proposed TeaNet.} \Circled{U} stands for feature concatenation.}
\label{fig:inpainting_unet_details}
\end{figure}

The reconstruction module creates artificial spectra for domain-specific data augmentation. It inputs a randomly masked real spectrum and outputs a full spectrum, by filling in the missing parts.  As the input and the output of this module are both spectra (1D signal), the reconstruction task falls into the category of dense regression where U-Net and its variants are shown to be the most effective architectures\cite{UNet-ronneberger2015u,UNet-liu2018image,UNet-zhou2018unet++,Liu2021Sundial,GUnetZhun2021}. We therefore \textcolor{black}{employ a U-Net architecture} in the reconstruction module as shown in Fig.~\ref{fig:inpainting_unet_details}. The reconstruction network contains 34 trainable convolutional layers in total and is larger than almost all neural networks previously used for vibrational spectrum recognition. 

\subsection{Training Protocol}
{\color{black}
The classification network is trained on real samples along with the corresponding reconstructed samples using the cross-entropy loss. }
The reconstruction network is trained to minimize the weighted sum of the classification (cross-entropy) and reconstruction (MSE) losses. 
The reconstruction network does not input class label of the sample, but the reconstruction is guided by the classification loss, meaning that the reconstructed sample must lie within the class boundaries, but otherwise could be arbitrary. In other words, the classification loss allows the reconstruction to move away from the original sample towards samples that minimize the classification loss.  To increase diversity in samples generated from the same input, it is essential to train both networks jointly. If the reconstruction loss dominates the training process of the reconstruction module, it might quickly converge and learn to repair a masked spectrum perfectly. Then, the spectra that it generates would have no/little difference from the original ones and play no role in data augmentation. The algorithm for training TeaNet is detailed in Algorithm~\ref{alg:teanet}.

{\color{black}
 An important question is how many samples to generate from the original sample. On the one hand, generating lots of original-masked pairs allows us to exploit the training samples in a novel way for training a larger model that may help achieving better accuracy in the case of few-shot spectrum recognition. On the other hand, as the number of pairs grows, the gain from training on those pairs might peak at some point and then start decreasing. We verified this behaviour in our experiments (see Table ~\ref{Tab:AblaJointOrSep}). The extra information/variation embedded in these pairs is bounded, thus many of them will become more and more similar as more masked spectra are generated.

 \begin{algorithm}
\label{alg:teanet}
\SetAlgoLined
\justifying{
\noindent
{\textbf{Input}:
A training set $S_{t} =  \{(x_{i}, l_{i}) ,i=1,\cdots,N \}$ and a masked set  $\{(\bar{x}_{i}, l_{i}) ,i=1,\cdots,N \}$ generated from $S_{t}$. A validation set $S_{v}$ for selecting the best model. The reconstruction and classification modules $\Phi_{\Theta_{R}}$, ${D}_{\Theta_{C}}$ with trainable weights $\Theta_{R}$ and $\Theta_{C}$. $\lambda_{1}, \lambda_{2}$ are step sizes. $\gamma \in (0,1)$ is predefined to balance the reconstruction and the classification loss.}
}

\noindent
\For{t = 0 to T} 
{
\noindent
\textbf{(1)} \textcolor{black}{The reconstruction module takes a mini-batch of masked samples $\{\bar{x}_{i},i=1,\cdots,m\}$ and outputs the repaired ones $\{\widehat{x}_{i}^{(t)},i=1,\cdots,m\}$}, i.e. 
    \[  {\widehat{x}_{i}}^{(t)} = \Phi_{\Theta_{R}^{(t)}}(\bar{x}_{i}) \]
The reconstruction loss ${L}_{R}^{(t)}$ is computed as
    \[ {L}_{R}^{(t)} = \frac {1}{m} \sum_{i=1}^{m}\big\|{\widehat x}_{i}^{(t)} - x_{i} \big\|_2 \] 
where $m$ is the size of a mini-batch.

\noindent
\textbf{(2)} The classification module takes both the original and repaired samples $\{ x_{i},  \textcolor{black}{\widehat{x}_{i}^{(t)}},i=1,\cdots,m\}$, denoted as \textcolor{black}{$\widetilde{x}_{i}^{(t)}$} to simplify the notation, and predicts their labels 
\[ {\widetilde{y}_{i}}^{(t)} = {D}_{\Theta_{C}^{(t)}}({\widetilde{x}_{i}}^{(t)})\]
The cross-entropy loss \textcolor{black}{$L_{C}^{(t)}$} for classification is computed as 
    \[ {L}_{C}^{(t)} = -\frac {1}{m} \sum_{i=1}^{m} \mathcal{I}({\widetilde y}^{(t)}_{i} = l_{i}) \cdot \ln {\widetilde y}^{(t)}_{i}\]     
where $\mathcal{I}$ is an indicator function.

\noindent
\textbf{(3)} Calculating the joint-loss \textcolor{black}{$L_{J}^{(t)}$} 
    \[ L_{J}^{(t)} = \gamma * L_{R}^{(t)} + (1 - \gamma) * L_{C}^{(t)} \]
and updating ${\Theta_{R}}$ and ${\Theta_{C}}$
    \[ {\Theta_{R}}^{(t + 1)} = {\Theta_{R}}^{(t)} - \lambda_{1} \frac{\partial L_{J}^{(t)}}{\partial \Theta_{R}}\]
    \[ {\Theta_{C}}^{(t + 1)} = {\Theta_{C}}^{(t)} - \lambda_{2} \frac{\partial L_{C}^{(t)}}{\partial \Theta_{C}}\]
 }
 \noindent
 \KwOut{$D^{*}_{\Theta_{C}}$ is the classification module with the smallest \textcolor{black}{$L_{C}^{(t)}$} on the validation set $S_{v}$.}
\caption{TeaNet: Task-Enhanced Augmentation Network}
\end{algorithm}

\subsection{Masking Schemes}\label{Sec:Subsection_masking_schemes}

Generation of masks is a key component of the proposed approach. Fig.~\ref{Fig:masked_examples} shows examples of masked spectra and the corresponding original ones.
There are several aspects  we need to consider in designing the masking scheme. Specifically,  should the masks be generated randomly or with fixed position and size and how large they should be?  

In general any mask can be represented as a union of small masks $\mathbf{U}\{(a_{i},b_{i}), i=1,\cdots, u \}$ where $a_{i} \in [0,1),b_{i} \in (0,1]$ denote the relative position of the ends of each segment, and $u$ is the number of segments.

For convenience, we define another variable $\tau$, which is the ratio of mask to spectrum length
\begin{align}
    \tau = \Sigma_{i=1}^{u}  |b_{i} - a_{i}| 
    \label{Equ:mask_spectrum_ratio_tau}
\end{align}
Clearly $\tau \in (0, 1)$. As $\tau$ becomes larger, the difficulty of predicting the masked parts of a spectrum increases, requiring the embedding to capture the semantics of the domain and not only trivial dependencies. However, when the mask is too wide, the reconstruction becomes ambiguous which could slow down the learning process. 

We consider the following masking schemes controlled by the parameters $a$ and $\tau$:
\textit{\begin{itemize}
    \item MS1: both $a$ and $\tau$ are fixed
    \item MS2: $a$ is fixed, $\tau$ is random
    \item MS3: $a$ is random, $\tau$ is fixed
    \item MS4: both $a$ and $\tau$ are random
    \item MS5: union of several masks of MS4 
\end{itemize}}

{\textcolor{black}{
Our designs of masking schemes include fixed masks (MS1), random masks (MS4 and MS5) and hybrid masks (MS2 and MS3). All schemes, except for MS5,  generate continuous masks.  MS5 might produce several masked fragments that are disconnected from each other. These designs were inspired by masking schemes in existing work~\cite{zhong2020random,yun2019cutmix,Devlin2019BERTPO} and inpainting tasks in computer vision~\cite{InpaintingReview2020, Pathak2016, UNet-liu2018image}. In particular, the masking scheme MS4 (random masking) can be regarded as a specific version of Random Erasing (with fixed fillings of zeros) within the spectrum context, except for some specific design choices.} Our experiments in Section 5.5.1 showed that {\textcolor{black}{reconstructing with} MS4 with $a \in \mathcal{U}(0, 1)$ and $\tau \sim \mathcal{U}(0.3, 0.7)$ (were $\mathcal{U}$ stands for uniform distribution) produced the best results.

\section{Experiment I: On Synthetic Spectra}\label{sec_experiment_onsyntheticspectra}

{\color{black}
To analyse the advantage of the proposed TeaNet compared to previous architectures (e.g., CNN~\cite{liu2017deep,LIU2019Dynamic}), we perform a set of tests on synthetic infrared spectra in which we can control the variation in data, in particular the location and appearance of the peaks (which are the most meaningful parts of a spectrum). 

The following experiments showed \textcolor{black}{that  TeaNet} is able to detect peaks in a wide range of scales. Namely, it is capable of capturing information from both sharp and wide peaks. \textcolor{black}{Conversely,} the performance of CNN depended on the scale of the \textcolor{black}{peaks}. It performed well on inputs in which characteristic peaks were sharper and much stronger than non-characteristic peaks. However its performance dropped significantly on spectra with wide and weak peaks. 

\begin{figure}
	\centering
	\includegraphics[width=\columnwidth]{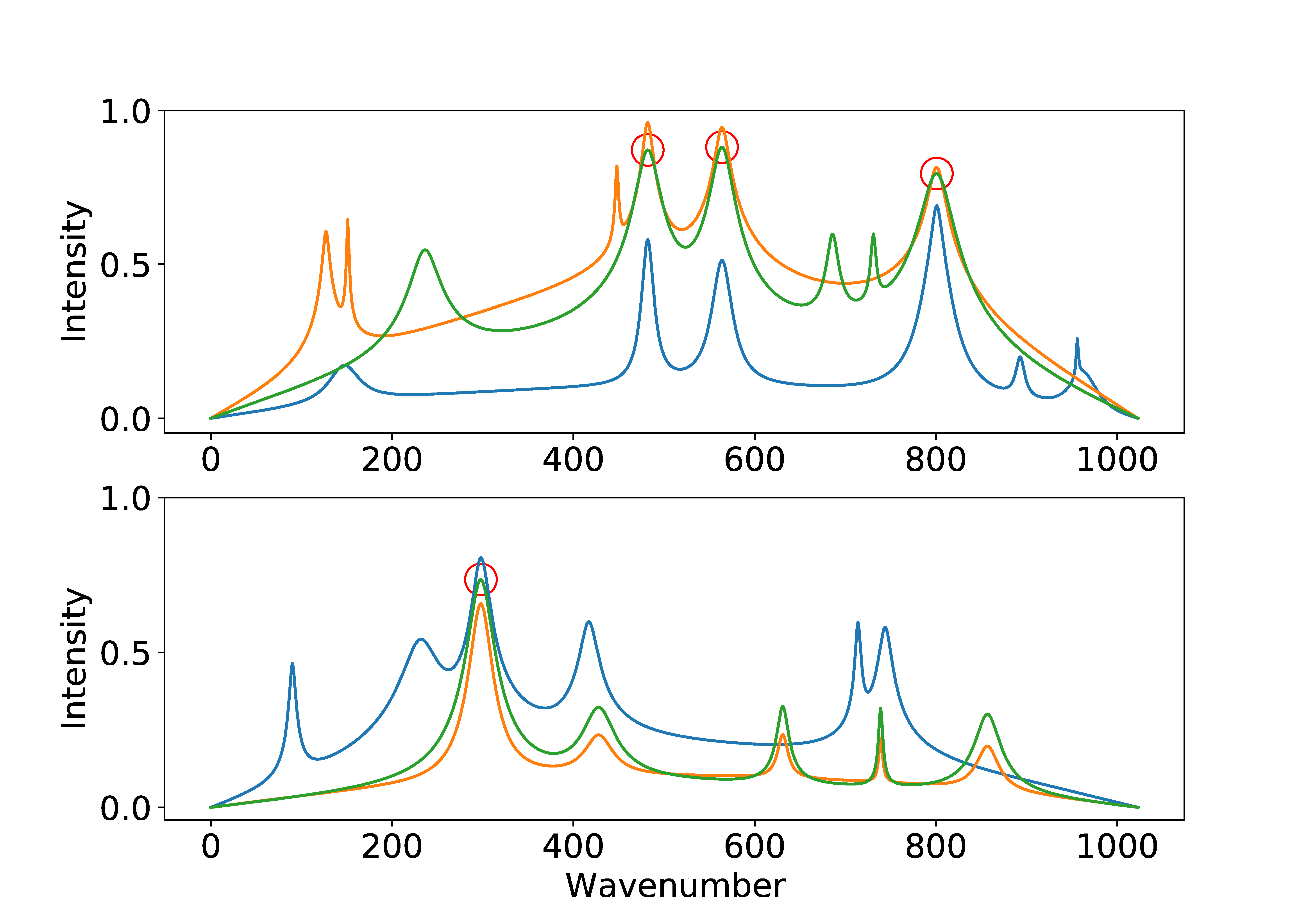}
    \caption{Synthetic spectra of two classes for few-shot classification created using the Lorentz model. Three samples of each class were plotted here with different colors. The characteristic peaks are marked by circles, while the rest are non-characteristic peaks or simply noise. For these samples, we have set $\vartheta=\frac{10}{3}$ and $\text{FWHM} \in 0 \sim 50$. Details can be found in section \ref{subsubsec_syntheticfewshot}.}
    \label{Fig:SyntheticSpectra_Examples_10vs3}
\end{figure}
}

\subsection{Generating synthetic spectra}
\subsubsection{Physical model}
We generate the synthetic spectra using the Lorentz model~\cite{Fukuhara2019} (in  Eq.~(\ref{Eq:Lorentz})), which is an infrared spectrum model used in physics:
\begin{align}
        f(p)=C+\sum_{\{\text{FWHM},A,p_{0}\}}{\frac{A}{1+\{{\frac{p-p_0}{\text{FWHM}/2}\}^2}}}
        \label{Eq:Lorentz}
\end{align}
where ${p}$ denotes a position/wavenumber, ${p_0}$ stands for the position of the maximum,  FWHM is the full width at half maximum, and $A$ controls the height of a peak. ${C}$ is a constant, which has been set to be zero in our experiments.

\subsubsection{Few-shot Classification Problem}\label{subsubsec_syntheticfewshot}
We set up the few-shot classification with $N_{s}$ ways and  $M_{s}$ shots, namely each classification problem included $N_{s}$ classes with $M_{s}$ samples in each. 
For each class, we first generated $K_{s}^{1}$ characteristic peaks at random. These peaks were shared among all samples in a class. We then produced $K_{s}^{2}$ non-characteristic (noisy) peaks with random or predefined parameters in Eq. (\ref{Eq:Lorentz}) and combined them with the characteristic peaks to form samples of the class. We set $N_{s} = 100$, $M_{s} = 20$, $K_{s}^{1} \in [1,3]$, $K_{s}^{2} \in [3,5]$. The length \textcolor{black}{of each spectrum} was 1024. We then generated 20 experiments as explained below with a different set of 100 classes, each class including 20 samples.

\subsubsection{Experimental Setting}
We denote $\vartheta$ to be the ratio of the height of the characteristic and non-characteristic peaks:

\begin{align}
    \vartheta = \frac{A {\text{ of a characteristic peak}}}{A {\text{ of a non-characteristic peak}}} 
\end{align}
$\vartheta$ controls the difficulty of the problem, i.e., when $\vartheta$ is smaller, it is more difficult for learning models to capture the discriminant information. 

To thoroughly investigate the proposed method, we varied two main factors:  FWHM (the width) of the characteristic peaks (signals) and $\vartheta$. We  tested five typical values of $\vartheta$: 10, $10/3$, 1, 0.3, 0.1 and four values of FWHM from $[0, 50]$ to $[150, 200]$. $\text{FWHM} \in [h_{l}, h_{r}]$ means when generating the spectra, FWHM were randomly generated from the region $h_{l} \sim h_{r}$. Consequently, we performed $5 \times 4 = 20$ sets of experiments in total. 

Fig.~\ref{Fig:SyntheticSpectra_Examples_10vs3} shows examples of two (out of 100) synthetic classes, generated with $\vartheta=\frac{10}{3}$, $\text{FWHM} \in [0, 50]$. For each class, it shows three samples in different colors. The characteristic peaks are marked by circles, while the rest are non-characteristic peaks. 
For each the synthetic problem, we randomly partitioned the data set into training, validation and test sets in the ratio of 60\%, 20\%, 20\%.

{\color{black}
\subsection{Network Architecture and Training Protocol} \label{subsubsec_synthetic_network_arch}

For comparison, we trained a double-pyramid LeNet, denoted as CNN\_Full, for the synthetic classification problems, the detailed architecture of which can be found in Fig.~\ref{fig:classification_details}(a). 

For TeaNet, we have kept the reconstruction module for all the experiments presented in this paper, as shown in Fig.~\ref{fig:inpainting_unet_details}. The masking scheme was MS4 with $a \in \mathcal{U}(0, 1)$ and $\tau \sim \mathcal{U}(0.3, 0.7)$. The classification module is the same as the CNN\_Full. 

To train CNN\_Full, we used an Adam optimizer with \textcolor{black}{a learning rate of $1\mathrm{e}{-4}$ and a batch size of 256.
We also used Adam} optimizers for both reconstruction and classification modules \textcolor{black}{with learning rates of $1\mathrm{e}{-3}$ and $1\mathrm{e}{-4}$ respectively and a batch size of 256}. The maximum number of epochs was 300. All the experiments were carried out on multiple NVIDIA GTX-3090(3080) GPUs. These settings have also been adopted for experiments on real-world datasets.
}

\begin{table}[!h]
\centering
\caption{Performance comparison of TeaNet and CNN on experiments with synthetic spectra. }
\begin{threeparttable}
\begin{tabular}{lcc}
\toprule
 \diagbox{FWHM}{$\vartheta$} & large & medium/small \\
\midrule
small  &  TeaNet $\approx$ CNN & TeaNet $\approx$ CNN\\
large  & TeaNet $\approx$ CNN &  TeaNet $\gg$ CNN\\
\bottomrule
\end{tabular}
\begin{tablenotes}
\footnotesize
\item $A \approx B$ - the performance of A and B are similar. 
\item $A \gg B$ - A performs much better than B.
\end{tablenotes}
\end{threeparttable}
\label{Tab_SyntheticResultTable}
\end{table}

\subsection{Results and Analysis}

\begin{figure}[!htp]
	\centering
	\includegraphics[width=0.5\textwidth]{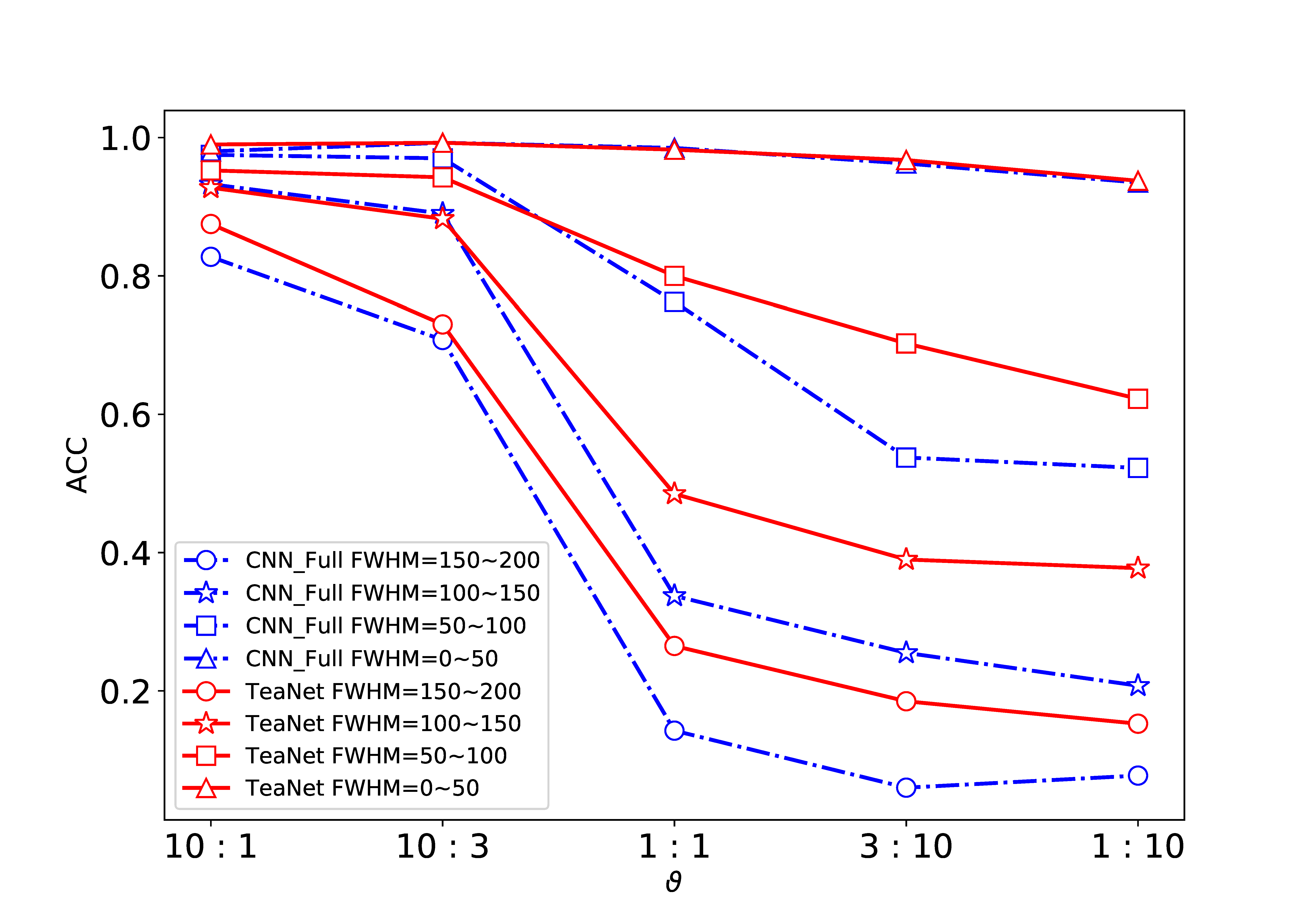}
    \caption{Comparison of TeaNet with CNN on a synthetic dataset with different settings of characteristic peaks(signals) and non-characteristic peaks(noise or interference). $\vartheta$ denotes the ratio of the height of the characteristic peaks and the non-characteristic peaks. Note that when $\vartheta$ is medium or small and FWHM is large, the spectrum recognition becomes very difficult. In this case, TeaNet outperformed CNN by a large margin. }
	\label{Fig:TeaNetvsCNN_OnSyntheticData}
\end{figure}

We investigated four typical scenarios shown in Table~\ref{Tab_SyntheticResultTable} and the results are shown in Fig.~\ref{Fig:TeaNetvsCNN_OnSyntheticData} and summarised in Table~\ref{Tab_SyntheticResultTable}. When $\vartheta$ is large, in which case the characteristic peaks dominate the non-characteristic ones, both methods performed equally well. As $\vartheta$ decreases, the recognition task become harder. In this setting, TeaNet performed much better than CNN by a large margin, though the accuracy of both methods dropped.

\section{Experiment II: On Real-world Spectra}\label{sec:experiments_realdata}
\subsection{The Compared Methods}
We compared the proposed TeaNet to several alternative augmentation schemes, namely,  1) plain augmentation: CNN\_Full~\cite{liu2017deep}, 2) inputting partial spectra by randomly masking the full spectra: CNN\_Partial, \textcolor{black}{3) randomly masking followed by filling with random patches from other samples: CutMix~\cite{yun2019cutmix}, 4) randomly combining samples linearly: MixUp~\cite{zhang2017mixup}}, 5) generating new samples via GANs: CWGAN~\cite{luo2018eeg}, DAGAN~\cite{antoniou2017data}, 6) augmentation in the feature domain: LTSA~\cite{jenni2018self}. 

CNN\_Full was included as the main baseline. We included CNN\_Partial for two reasons. First, partial images, aka patches, have been used in computer vision for data augmentation to facilitate the training of large CNNs. Training the classifier on masked spectra is equivalent to training on image patches. The second reason is more hypothetical. Due to randomness, there is a chance that some of the masked spectra would contain only the characteristic peaks. In that ideal case, the masking process would be equivalent to a randomized wavenumber selection, which would lead to a performance improvement. In fact, in real world applications, especially in industry, it is a common practice to use human experts for manually selecting the discriminant wavenumbers and then using them as an input to the conventional machine learning methods, such as PLS. However, in reality, many of the randomly masked spectra would contain lots of non-characteristic peaks, and would require \textcolor{black}{the} CNN to detect the characteristic peaks in the masked samples and ignore the noise. So, we may view CNN\_Partial as a crude attempt of realizing an automatic wavenumber selection procedure. 

\textcolor{black}{
CutMix and MixUp are both popular augmentation methods used in computer vision and to some extent both are related to masking for augmentation. In the context of spectrum recognition, CutMix masks a spectrum randomly and then fills the masked region with random patches from other \textcolor{black}{spectra}. MixUp combines multiple spectra linearly and conceptually may be regarded as a merging of masking and filling in one step.
}
The rest of the methods ( GANs and LTSA)  are the natural competitors in data augmentation. Moreover, As CWGAN, DAGAN and LTSA were  originally proposed for images, we re-implemented them  with necessary modifications associated with 1D signals.

We also tested conventional machine learning methods, such as correlation, kNN, random forest, and  support vector machines, as shallow models usually require \textcolor{black}{fewer} training examples than the deep models. Moreover, these models are popular among practitioners.

\begin{table}[!h]
\centering
\caption{Benchmark datasets used in this study}
\begin{tabular}{lcccc}
\toprule
Dataset    & Type & \# Samples & \# Classes & \# Samples/class\\
\midrule
RRUFF\_IR  & IR  & 698   & 430 & $\approx$ 1.62\\
USGS  & NIR  & 887   & 212  & $\approx$ 4.18\\
ReLab  & NIR  & \textcolor{black}{6675}   & 855 & $\approx$ \textcolor{black}{7.81} \\
\bottomrule
\end{tabular}
\label{Tab:Dataset}
\end{table}

\subsection{Datasets and Evaluation Protocol}\label{Subsec_evalprotocols}
In our experiments, we used three publicly available datasets, namely RRUFF\_IR~\cite{lafuente20151_rruff}, USGS~\cite{kokaly2017usgs}, ReLab. All these datasets contain infrared spectra of minerals. Details can be found in Table \ref{Tab:Dataset}. It is worth noting that these  datasets contain 1.62, 4.18 and 7.81 samples per class on average, which fit the few-shot learning setting. 

As the number of samples per class is rather small (see Table~\ref{Tab:Dataset}), we performed the leave-one-out scheme to generate a test set with a single sample per class (chosen at random). The rest of the samples were used for training and validation. We repeated this procedure $N=10$ times and we reported the averaged accuracy over these experiments. 

\subsection{Network Architecture and Training Protocol}
For the classification module in TeaNet, we adopt a double-pyramid LeNet with the number of convolutional blocks optimised for each benchmark data set. For USGS, the classification module was the same as the one described in~\ref{subsubsec_synthetic_network_arch}, as shown in Fig.~\ref{fig:classification_details}(a). For the RRUFF\_IR and Relab datasets, the classification module included two convolutional blocks with one dense layer. Details can be found in Fig.~\ref{fig:classification_details}(b). 
The architecture of the reconstruction module in TeaNet was the same in all experiments. Please refer to Fig.~\ref{fig:inpainting_unet_details} for more details.

We used Adam optimizer for both reconstruction and classification modules with the learning rate: $1\mathrm{e}{-3}$ and $1\mathrm{e}{-4}$ respectively and 256 batch size. The maximum number of epochs was 300. All the experiments were carried out on multiple NVIDIA GTX-3090(3080) GPUs.

\subsection{Results and Analysis}
\textcolor{black}{
\subsubsection{Performance on benchmark datasets}}
Results in Table \ref{Tab:Main_Benchmark_Results} show the classification accuracy of the compared methods on the benchmark datasets. First, it is evident that the proposed TeaNet significantly outperforms all the compared methods achieving the state-of-the-art results on all three datasets.

The results of the conventional machine learning methods, such as correlation, kNN, random forest and support vector machines show that  even though the conventional methods generally require less data than the deep  \textcolor{black}{models, their accuracy is poorer} as they are unable to recover useful features and thus require hand-crafted feature engineering. 

The accuracy of LTSA on all three datasets is worse than that of the other deep networks by a large margin. In particular, its accuracy is on average nearly 10\% lower than the baseline CNN\_Full. We \textcolor{black}{believe} that even though LTSA utilized inpainting -- a concept similar to ours, it was not designed to deal with the few-shot learning scenarios, but rather to learn better representations with the help of pretraining.

\begin{table}
\centering
\caption{Classification accuracy (\%) of the compared methods on several benchmark spectral datasets}
\begin{threeparttable}
\begin{tabular}{lcccc}
\toprule
\diagbox{Method}{Dataset}& RRUFF\_IR       & USGS            & ReLab               \\
\midrule
kNN          & 79.32\%$\pm$1.56 & 64.73\%$\pm$2.31 & 73.33\%$\pm$0.87   \\
SVM(linear)     & 71.22\%$\pm$1.29 & 50.70\%$\pm$2.00 & 29.91\%$\pm$1.13   \\
SVM(rbf)        & 76.33\%$\pm$1.17 & 62.17\%$\pm$2.21 & 49.52\%$\pm$1.13   \\
Random Forest   & 69.46\%$\pm$3.11 & 67.44\%$\pm$1.90 & 70.68\%$\pm$0.59   \\
Correlation     & 74.69\%$\pm$1.97 & 38.14\%$\pm$1.54 & 51.95\%$\pm$0.84   \\
\midrule
LTSA~\cite{jenni2018self}        & 80.98\%$\pm$2.60  & 72.94\%$\pm$3.44  & 74.66\%$\pm$4.04  \\
CWGAN~\cite{luo2018eeg}          & 81.97\%$\pm$2.13  & 82.56\%$\pm$3.31  & 82.47\%$\pm$0.99  \\
DAGAN~\cite{antoniou2017data}    & 88.98\%$\pm$1.21  & 83.57\%$\pm$1.42  & 82.97\%$\pm$0.92  \\
\midrule
\textcolor{black}{MixUp~\cite{zhang2017mixup}}  & \textcolor{black}{88.85\%$\pm$1.56}  & \textcolor{black}{84.58\%$\pm$1.32} & \textcolor{black}{78.44\%$\pm$0.87} \\
\textcolor{black}{CutMix~\cite{yun2019cutmix}} & \textcolor{black}{87.21\%$\pm$1.49}  & \textcolor{black}{85.43\%$\pm$1.29}  & \textcolor{black}{82.84\%$\pm$0.98}   \\
CNN\_Full~\cite{liu2017deep} & 88.03\%$\pm$0.98  & 84.81\%$\pm$1.56  & 82.06\%$\pm$1.75  \\
CNN\_Partial      & 86.53\%$\pm$1.32  & 86.59\%$\pm$1.55  & 83.90\%$\pm$1.23  \\
\midrule
TeaNet (ours)    & {\color{black}\textbf{90.07}\%$\pm$1.33}   & {\color{black}\textbf{87.37}\%$\pm$1.47}  &  {\color{black}\textbf{85.43}\%$\pm$1.00} \\
\bottomrule
\end{tabular}
\end{threeparttable}
\label{Tab:Main_Benchmark_Results}
\end{table}

We note that CWGAN~\cite{luo2018eeg} is merely a plain GAN and serves as a baseline for GAN-like methods in our tests. DAGAN was specifically designed for data augmentation and thus was expected to perform better than the plain CWGAN. Note that on RRUFF\_IR and USGS, which have fewer samples per class, DAGAN performed better than CWGAN, yet still worse than the baseline CNN on full spectra. 
It is not surprising that a plain GAN failed to generate sensible samples when only few samples were provided for its training, as GAN is generally known as a data demanding model. 
When relatively more samples were available, as in ReLab dataset, both CWGAN and DAGAN performed better and the accuracy increased by about 0.4\% and 1.0\% respectively.

\textcolor{black}{While similar to approaches for training on image patches, which are beneficial in computer vision problems, CNN\_Partial did not perform well in all cases.} We observe that, training on partial/masked samples directly improved the accuracy by 1.8\% and 1.9\% on USGS and ReLab respectively. However, on RRUFF\_IR  its accuracy dropped dramatically by 1.5\%. 

{\color{black}
Recall that CutMix can be viewed as random masking followed by a naive/rough reconstruction, so if the naive reconstruction helps, one should observe a performance improvement over CNN\_Partial. This, however, was not \textcolor{black}{the case}. In fact, the accuracy on USGS and ReLab dropped by around 1\% . MixUp also experienced a significant performance drop on these two datasets, though it performed relatively well on RRUFF\_IR. Overall \textcolor{black}{both methods performed} significantly worse than the proposed TeaNet. This may indicate that, in the case of small data where true variations are limited, generating artificial samples with sensible variations is of greater importance, though that dropping/masking a certain amount of regions with a rough reconstruction could help regularize the model when a large amount of data is available.
}

\textcolor{black}{
\subsubsection{Analysis of augmented samples}}

\textcolor{black}{ The purpose of the reconstruction module in our approach is to produce dataset augmentation for training the classifier. Thus the best way to evaluate the reconstruction module is by testing whether using reconstructed/augmented samples in classifier training improves the classification accuracy.
The results in the previous section show that the augmented samples generated by TeaNet facilitated the training of the classifiers. Moreover, to better understand the behavior of TeaNet, we further investigate the variations (of augmented samples) generated by the reconstruction module.} 

\textcolor{black}{
\textit{Visualization of generated variations}:}
To illustrate the variations generated by TeaNet, we plot four augmented spectra for each of two minerals, as shown in Fig.~\ref{Fig:example_variations_generated}. \textcolor{black}{The dissimilarity measured by MSE between these artificial samples and the corresponding original samples were also calculated to be 0.74$\pm$0.81 and 3.39$\pm$2.02, the former of which corresponds to the sample shown in the left column of Fig.~\ref{Fig:example_variations_generated}, the latter of which corresponds to the sample in the right column.} It can be seen that TeaNet was indeed able to generate sensible variation within each masked region. By randomly masking different wavenumbers/regions, we generated augmented \textcolor{black}{samples} where variation occurred in the masked regions. \textcolor{black}{These} augmented samples helped TeaNet to identify the discriminant regions of a spectrum.

\begin{figure*}
	\centering
    \subfigure{
    \includegraphics[width=0.475\textwidth]{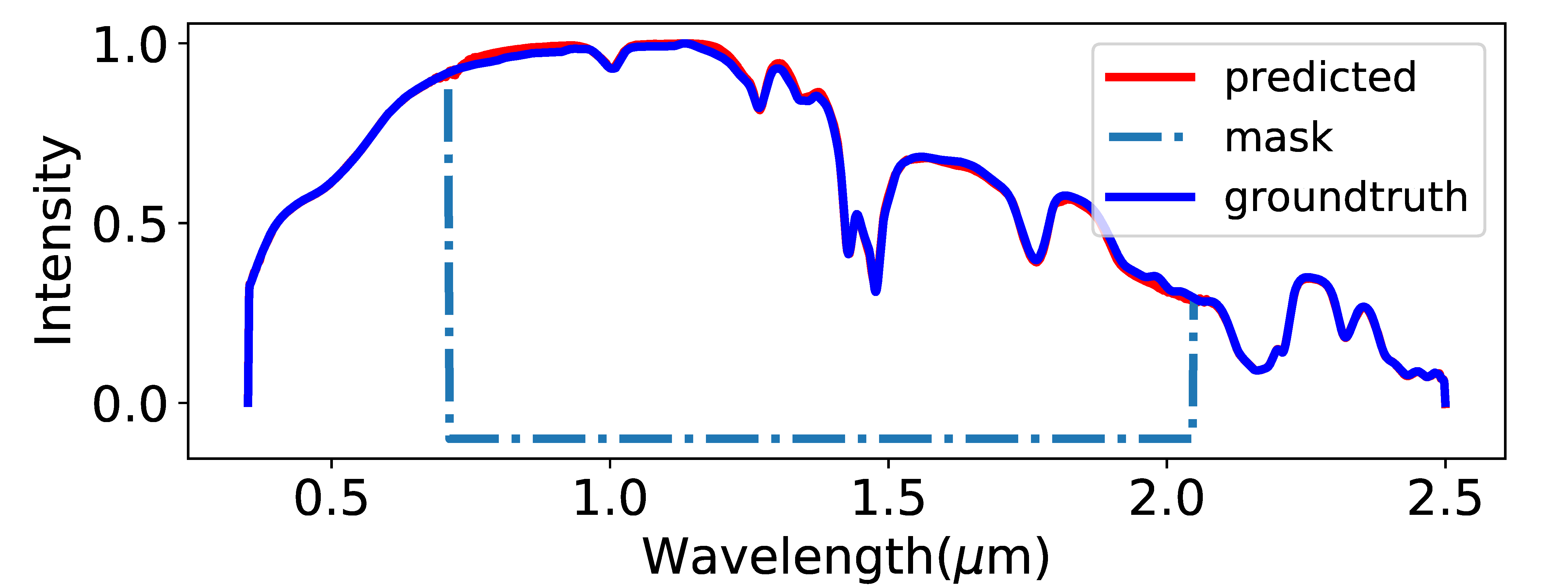}
    }
    \subfigure{
    \includegraphics[width=0.475\textwidth]{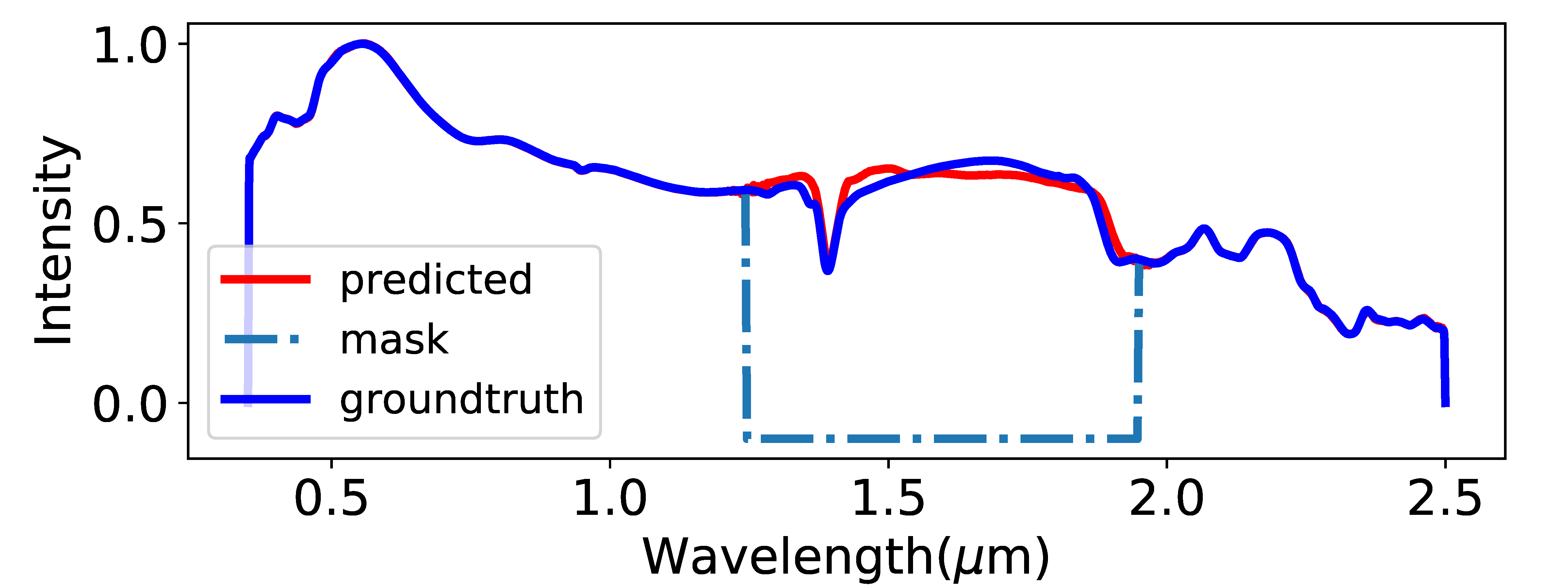}
    }
    \subfigure{
    \includegraphics[width=0.475\textwidth]{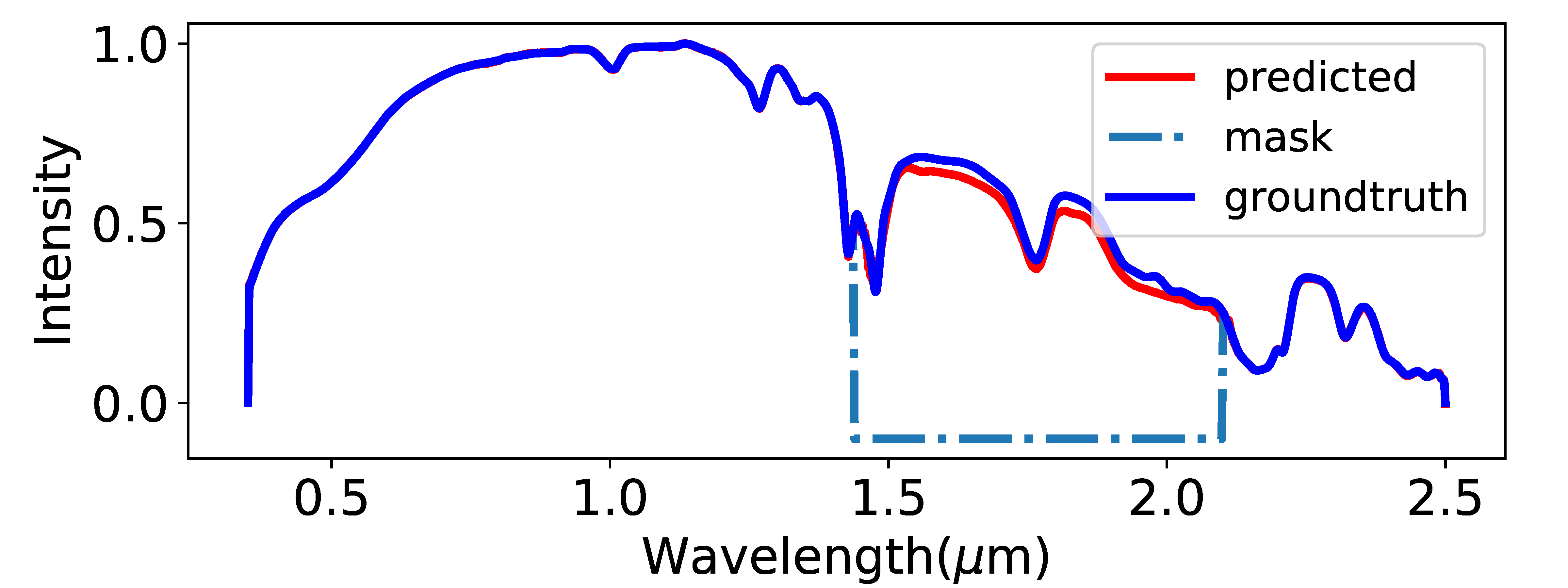}
    }
    \subfigure{
    \includegraphics[width=0.475\textwidth]{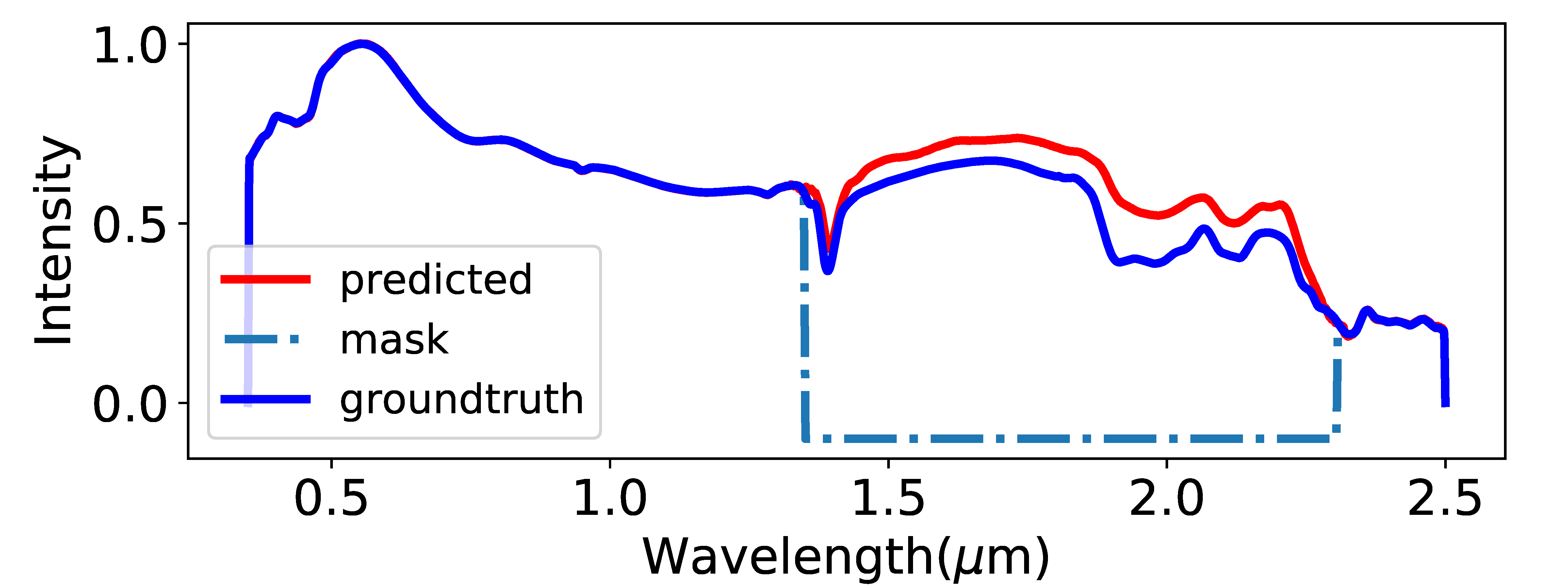}
    }
    \subfigure{
    \includegraphics[width=0.475\textwidth]{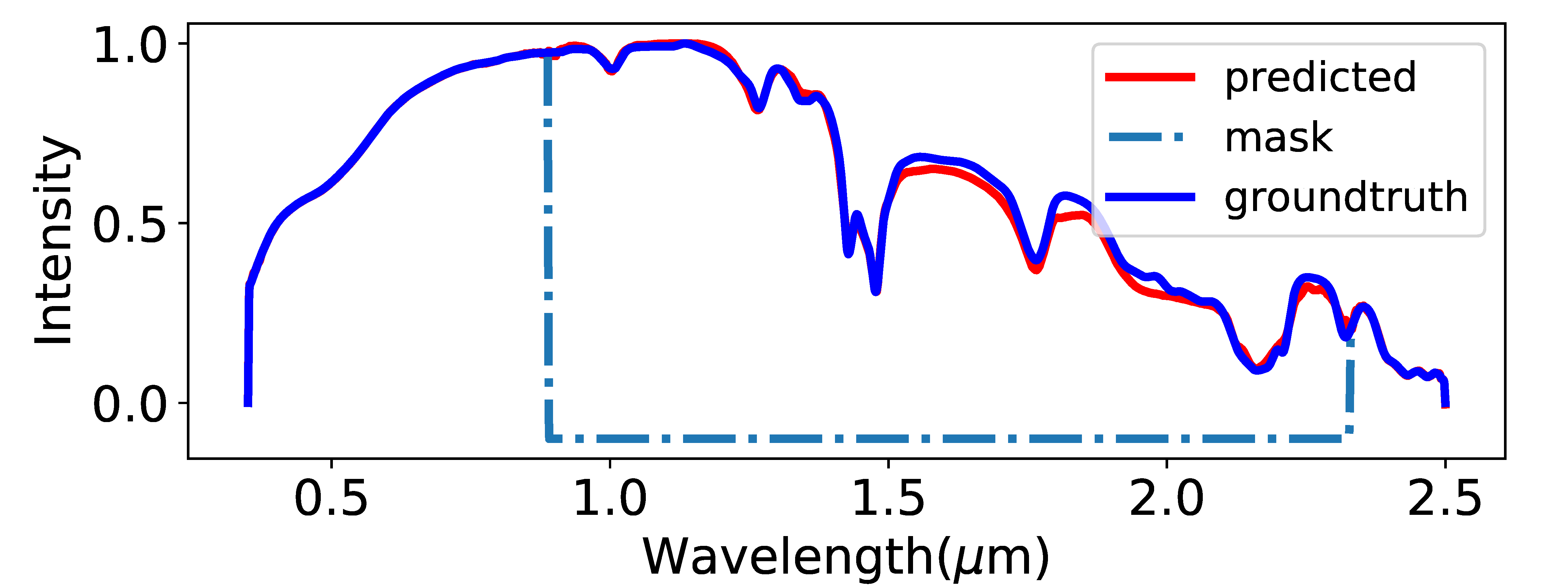}
    }
    \subfigure{
    \includegraphics[width=0.475\textwidth]{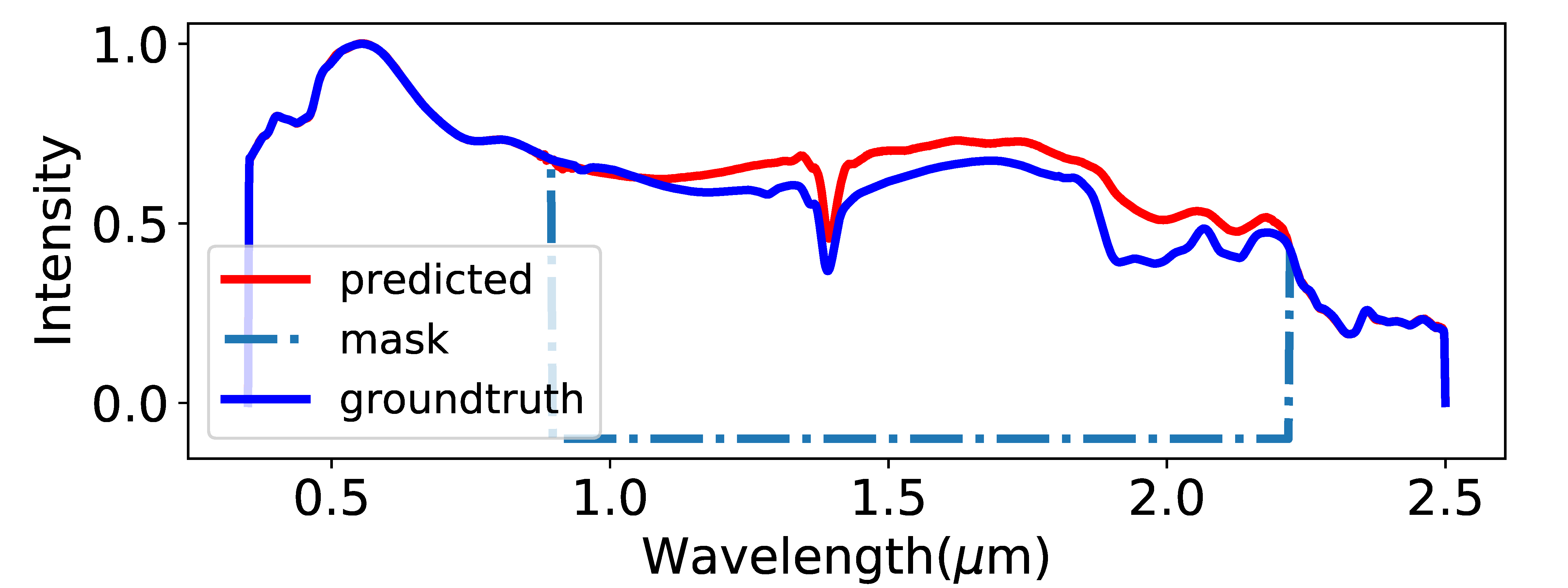}
    }
    \subfigure{
    \includegraphics[width=0.475\textwidth]{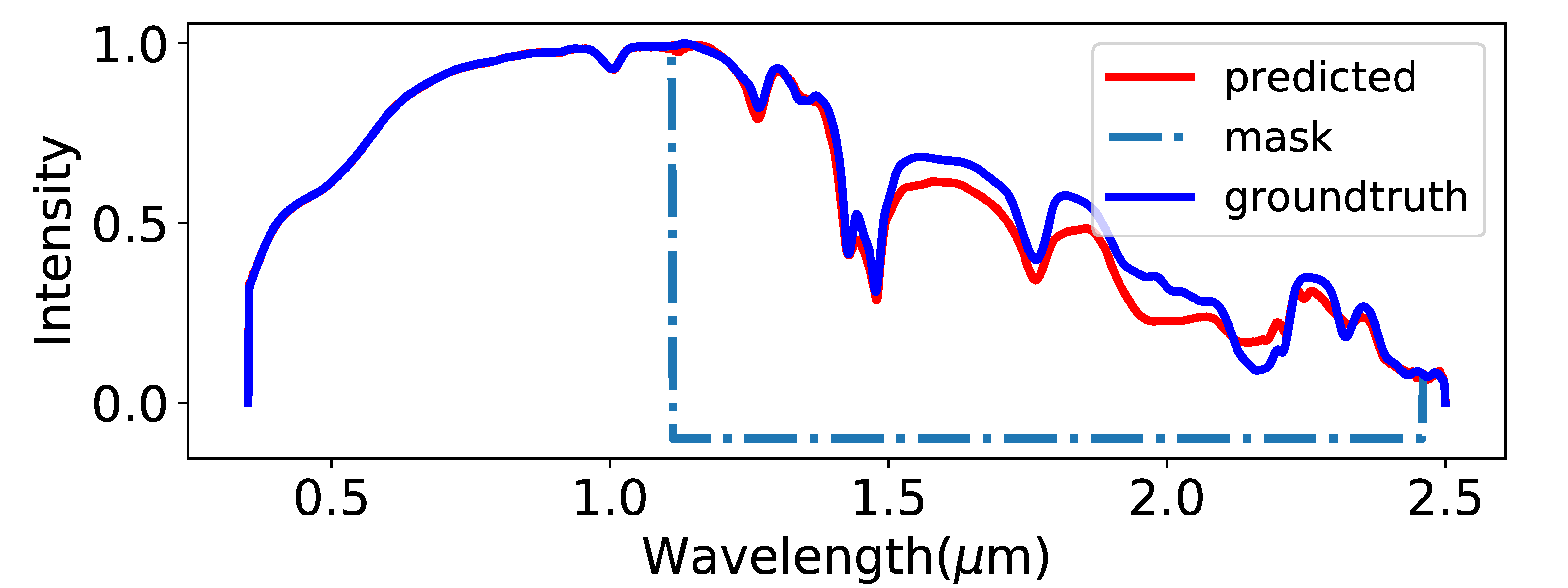}
    }
    \subfigure{
    \includegraphics[width=0.475\textwidth]{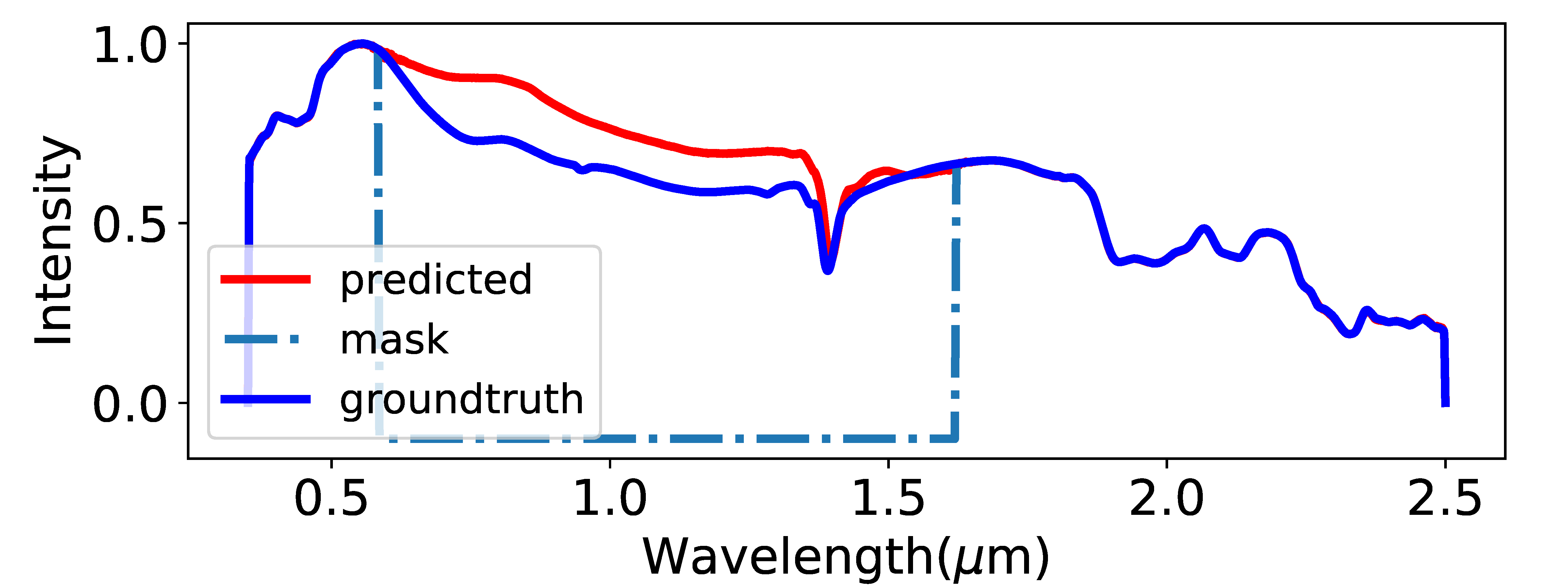}
    }
	\caption{Variations generated by predicting missing parts of spectra. The original spectra were plotted in blue. The masked regions were indicated by dot lines. The predicted spectra were highlighted in red. \textcolor{black}{The dissimilarity measured by MSE between these artificial samples and the corresponding original samples were calculated as 0.74$\pm$0.81 (left) and 3.39$\pm$2.02 (right)}. }
	\label{Fig:example_variations_generated}
\end{figure*}

\begin{figure*}
	\centering
    \subfigure[``Physical factors e.g. temperature changing'']{
    \includegraphics[width=0.475\textwidth]{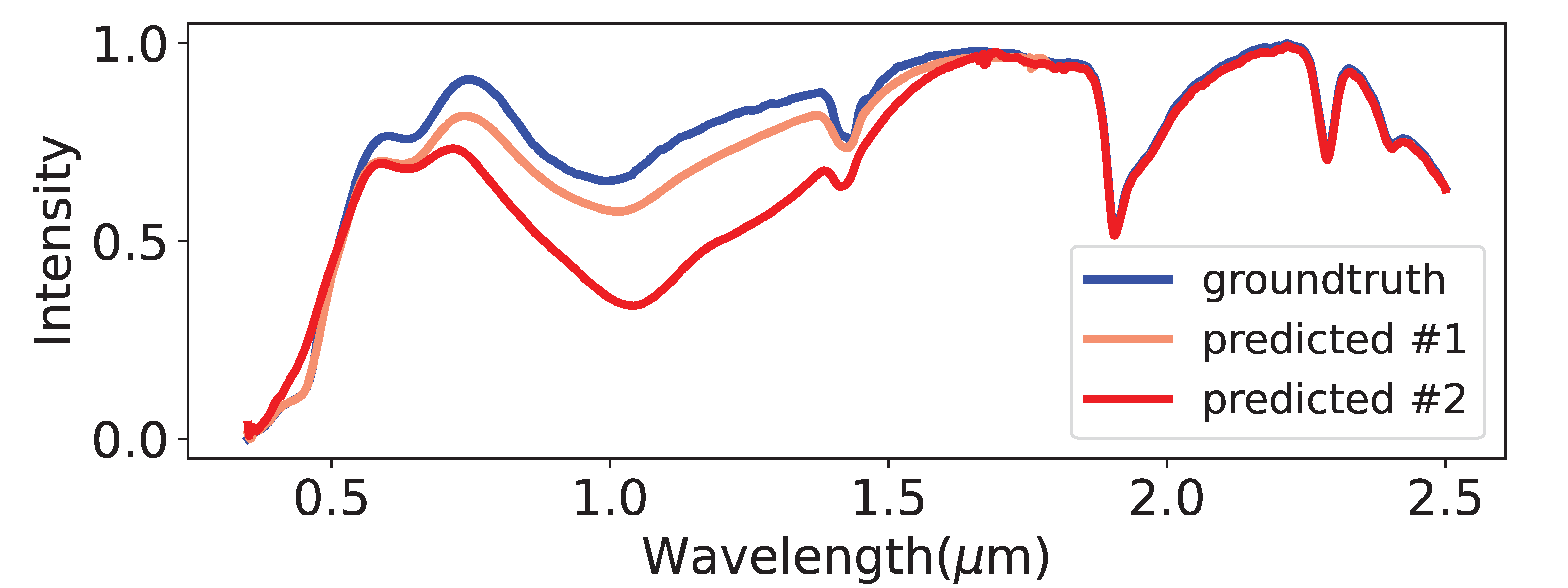}
    }
    \subfigure[``Smoothing'']{
    \includegraphics[width=0.475\textwidth]{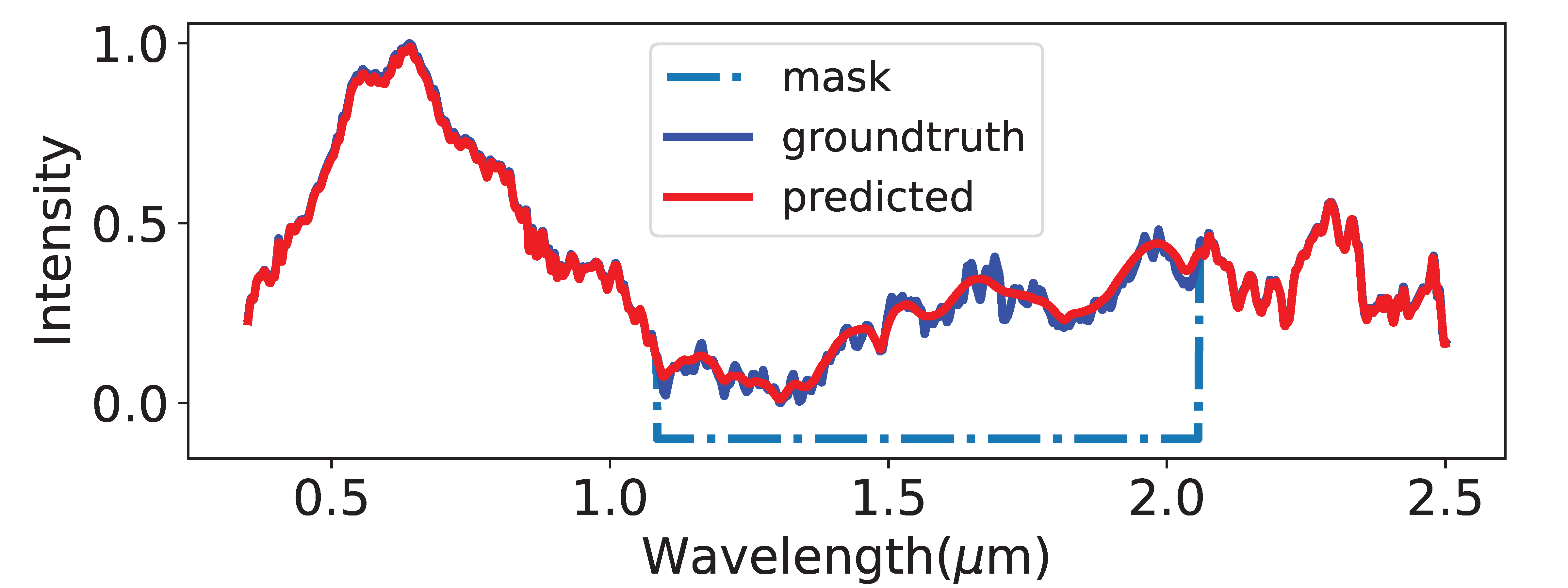}
    }
    \subfigure[``Baseline correction'']{
    \includegraphics[width=0.475\textwidth]{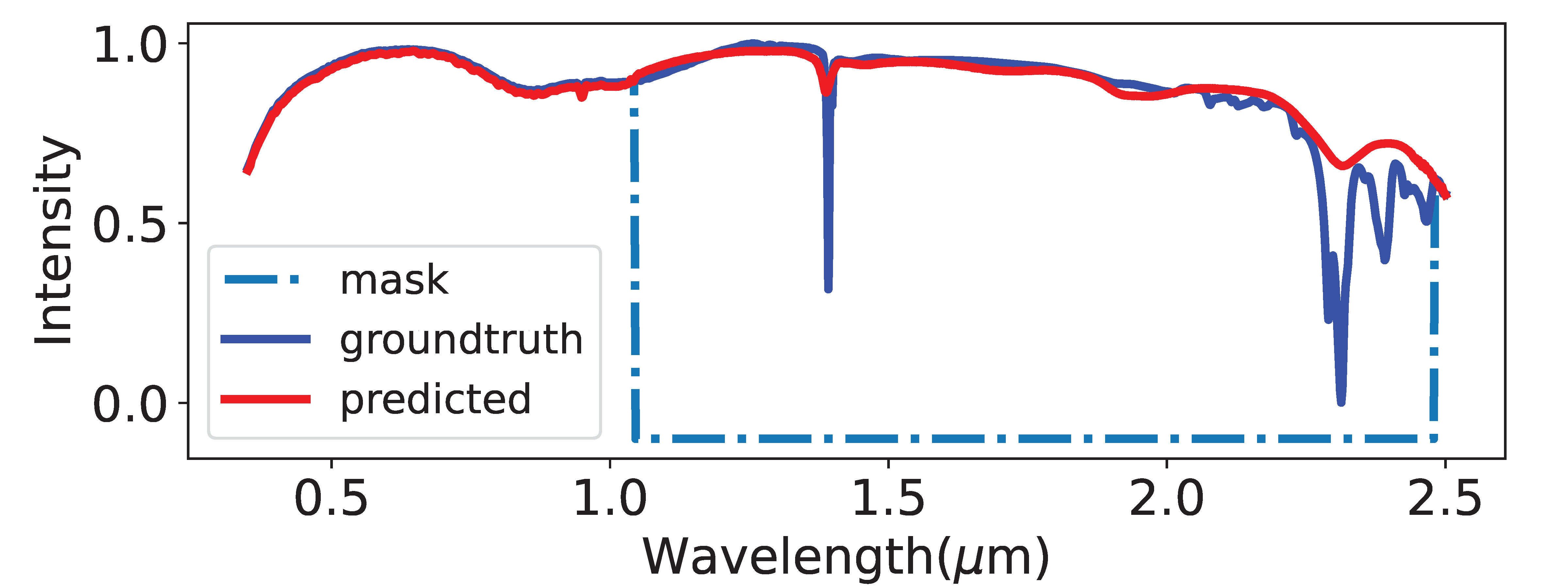}
    }
    \subfigure[``Masking out a peak'']{
    \includegraphics[width=0.475\textwidth]{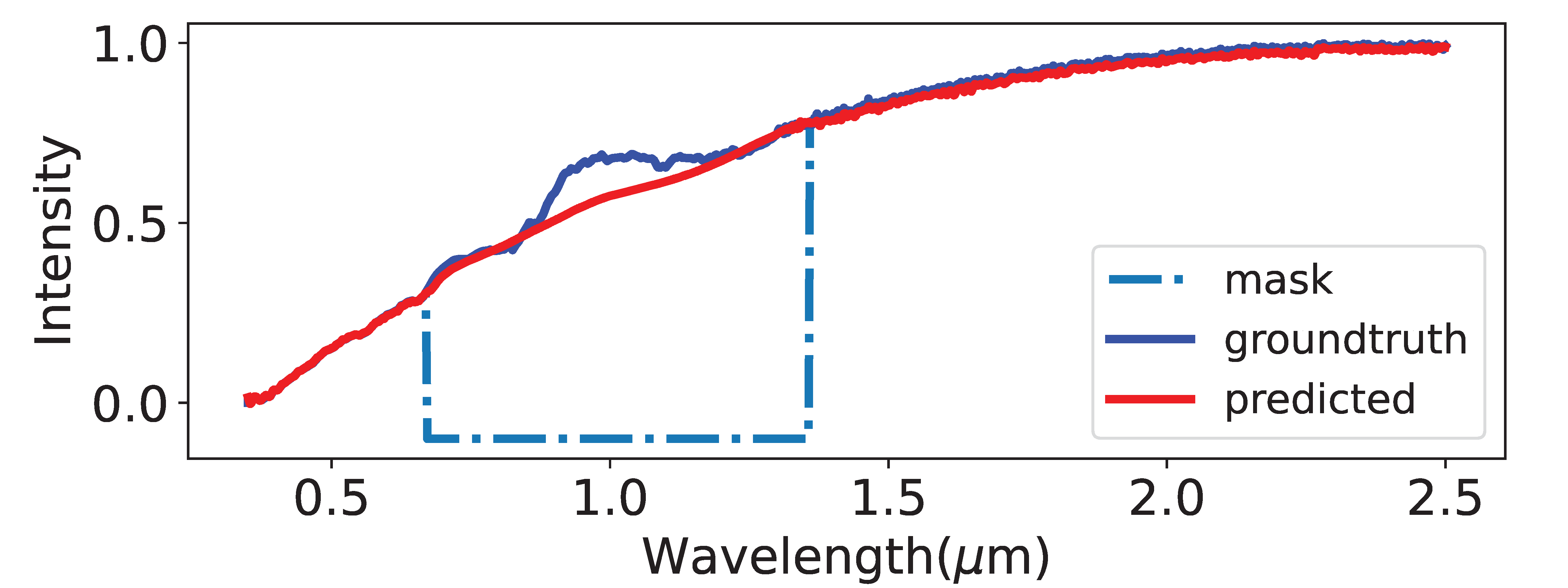}
    }
    \caption{\textcolor{black}{Typical examples of explainable variations generated by TeaNet. The original spectrum is plotted in blue. The augmented spectra are shown in red and pink. (a) {\color{black}Predicting different masked regions of the same spectrum produced variations which bear a resemblance to those induced by physical factors in the real world e.g. \textcolor{black}{changing temperature~\cite{FTIR_Weng}}. To avoid clutter, the masks are not shown.} (b) {\color{black}TeaNet generated a smoothed reconstruction of the masked region of a noisy spectrum.} (c) The augmented spectrum corresponds the baseline of the original one on the masked region. Here, TeaNet essentially performed a baseline correction which was a popular preprocessing technique for spectrum recognition when non-deep-learning methods were used~\cite{FTIR_Weng}. (d) TeaNet generated an augmented sample one peak of which was surgically removed.
    }}
	\label{Fig_example_variations}
\end{figure*}

{\color{black}
\textit{Explainability of generated variations}: 
It is not unexpected that augmented samples do not have to be human-understandable to assist the training of a complex deep neural network. A good related example is the popular method CutMix which has shown to be effective in image applications, though it generates incoherent artificial samples. On the other hand, we found that TeaNet was indeed able to generate explainable variations. To be specific, these variations generated by TeaNet not only obey the fundamental physical law of continuity, but also reflect physical factors or have performed distinct meanings or functions for learning.

Continuity is a fundamental physical law that a real spectrum should conform to. Though that augmented samples do not necessarily obey physical laws so as to facilitate the training of the classifiers,it is still of interest to investigate whether the augmented samples generated by TeaNet obey the physical law of continuity. It was observed in our experiments that all the generated samples were continuous and the reconstruction module learned to fill the masked region/wavenumbers without breaking the continuity. This suggested that the reconstructed spectra indeed conform to the physical law of continuity. In contrast, CutMix do not preserve the continuity as it simply fills the masked regions with random patches from other samples. 

Fig.~\ref{Fig_example_variations} shows four typical examples of explainable variations generated by TeaNet. 
Fig.~\ref{Fig_example_variations}(a) demonstrated that TeaNet was able to produce a series of augmented samples by reconstructing different masked regions of the same spectrum. The deformations/variations of these artificial samples bear a resemblance to those induced by physical factors in the real world e.g. temperature changing~\cite{FTIR_Weng}. Fig.~\ref{Fig_example_variations}(b) showed that TeaNet learned to generate a smooth reconstruction of the masked region of a noisy input. In Fig.~\ref{Fig_example_variations}(c), TeaNet generated the baseline of the input spectrum which implied that it essentially performed a baseline correction which was a popular preprocessing technique for spectrum recognition when non-deep-learning methods were employed~\cite{FTIR_Weng}. \textcolor{black}{ Fig.~\ref{Fig_example_variations}(d) shows an} augmented sample where one peak was surgically removed. 

Simulated variations which reflect physical factors can be beneficial to the training of the classifiers. \textcolor{black}{Both smoothing and baseline correction have been widely used as preprocessing for vibrational spectroscopy. Masking out a peak ``surgically'' may be viewed as a smart version of CutMix}. All these explainable variations demonstrated that TeaNet was indeed able to generate sensible augmented samples to assist the training of the classifiers. 
}

{\color{black}
\textit{\textcolor{black}{Cases for which reconstruction failed}}:
We observed that there were a small fraction, around $0.168\%$  on the ReLab dataset, of augmented samples with dramatic variations of large magnitude, the typical cases of which were shown in Fig.~\ref{Fig_example_failed_reconstruction}. \textcolor{black}{Purely from a reconstruction point} of view, these might be regarded as failure cases since the augmented samples were vastly different from the original ones. To be particular, the augmented sample shown in the first row of Fig.~\ref{Fig_example_failed_reconstruction} seems to be arbitrary, but after examination we found that it did not jeopardize the classification. We suspect that this large variation was generated probably because this particular class was easy to classify and therefore could tolerate large variations. 

Different from the first one, the variations shown in the second and third rows of Fig.~\ref{Fig_example_failed_reconstruction} were large, yet sensible, and could still be beneficial, because the corresponding augmented samples could be viewed as the original one with one peak/valley mirrored or trimmed respectively. In this sense, these samples might be meant to be produced to serve our main goal of generating artificial samples with sensible variations to help train the classifier.  

The averaged mask size of these samples with large variations was 0.58$\pm$0.11 where all the masks were sampled uniformly from 0.3 to 0.7. So we can reduce the number of the failure cases by imposing smaller masks if needed.\\
}

\begin{figure}[!h]
	\centering
    \subfigure{
    \includegraphics[width=0.475\textwidth]{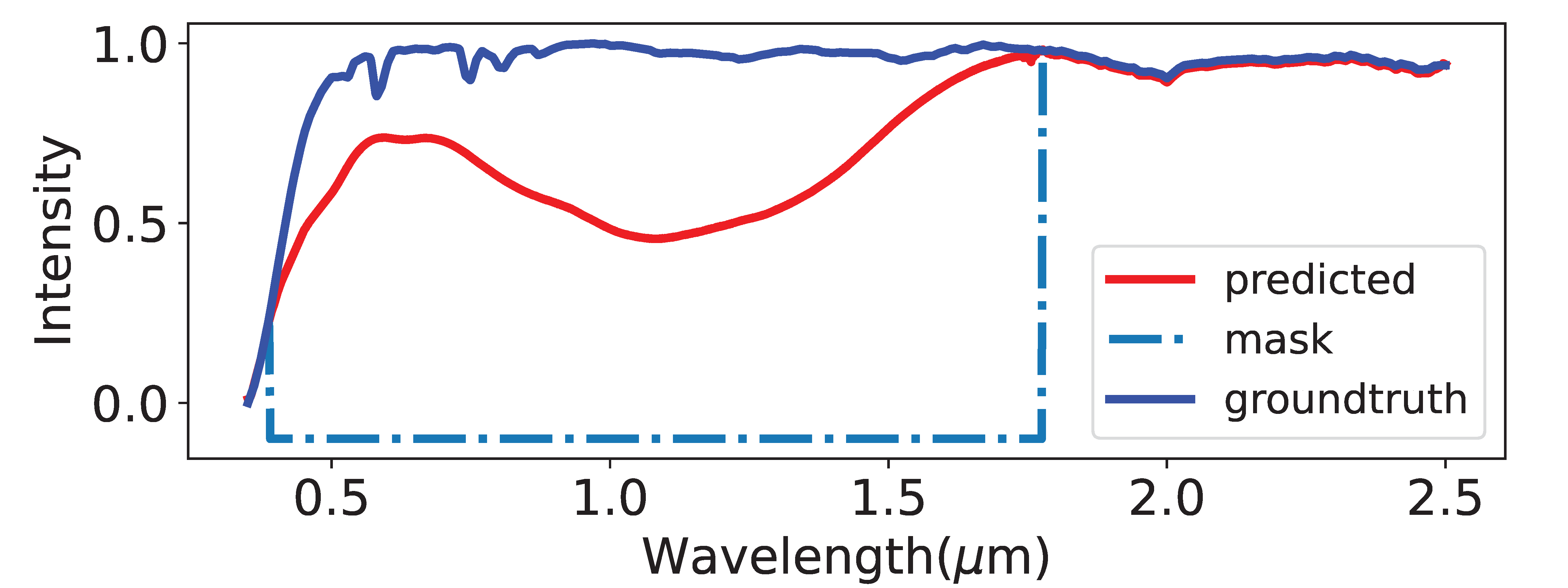}
    }
    \subfigure{
    \includegraphics[width=0.475\textwidth]{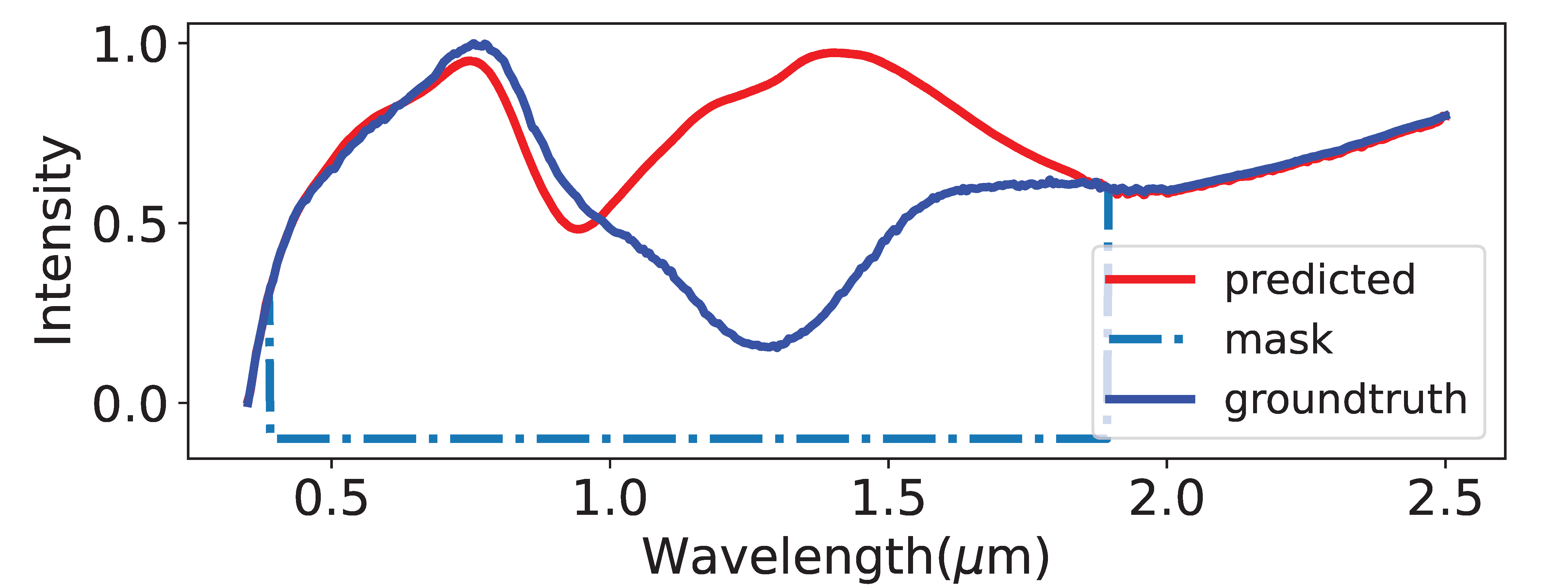}
    }
    \subfigure{
    \includegraphics[width=0.475\textwidth]{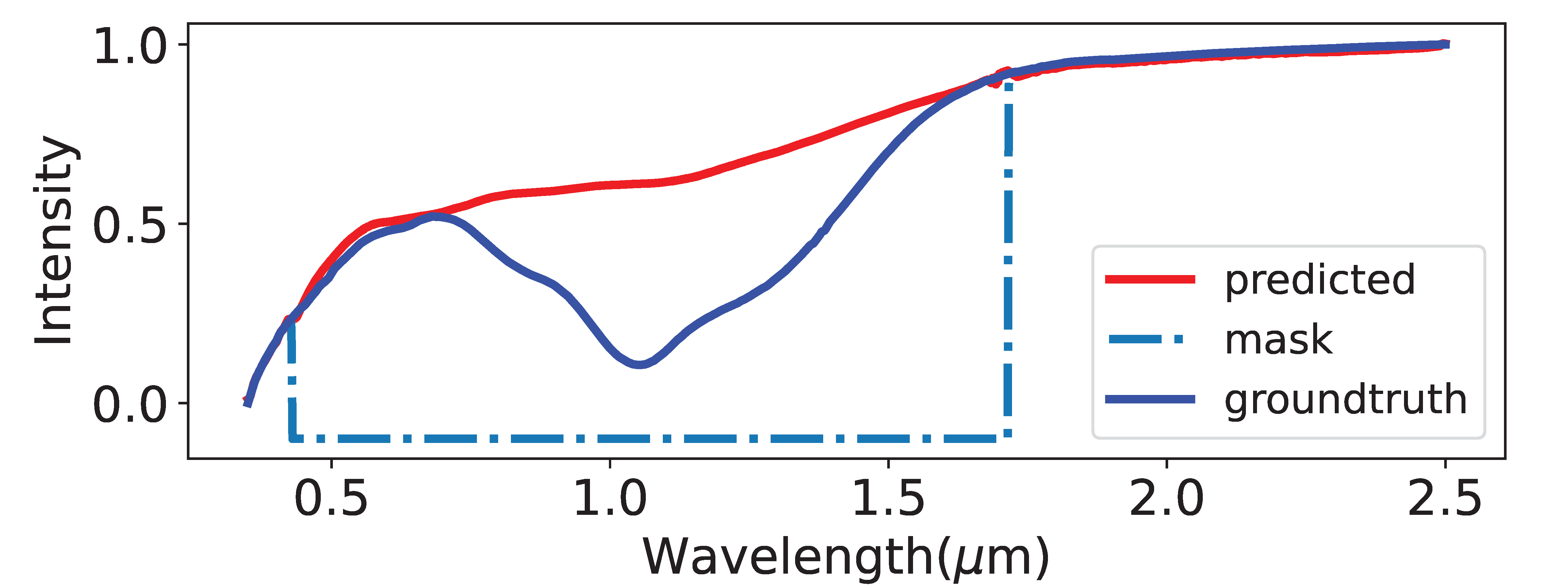}
    }
	\caption{\textcolor{black}{Large variations generated by TeaNet which might be considered as failure cases from reconstruction point of view as the reconstructed samples were greatly different from the original ones. The original spectra are plotted in blue. The masked regions are indicated by dot lines. The predicted spectra are highlighted in red.}}
	\label{Fig_example_failed_reconstruction}
\end{figure}

\subsection{Ablation Analysis}

\begin{table}
\centering
\caption{Comparison of different masking schemes on the benchmark spectral datasets}
\begin{threeparttable}
\begin{tabular}{lccc}
\toprule
\diagbox{Mask}{Dataset}& RRUFF\_IR       & USGS            & ReLab               \\
\midrule 
MS1: \\
$a=0, b=0.5$     & 87.35\%$\pm$1.86 & 82.71\%$\pm$2.30  &  80.89\%$\pm$1.51    \\
$a=0.5, b=1$     & 88.03\%$\pm$1.53 & 84.03\%$\pm$2.00  &  83.33\%$\pm$1.26    \\
\midrule
MS2: \\ 
$a=0, b\in (0, 1)$          & {\color{black}81.97\%$\pm$3.08} & 80.85\%$\pm$2.33 & 77.93\%$\pm$1.32 \\
$a=0.5, b\in (0.5, 1)$      & {\color{black}88.78\%$\pm$1.40} & 85.35\%$\pm$2.18 & 83.96\%$\pm$0.61 \\
\midrule
MS3: \\
{\color{black}$\tau$ = 0.01} & {\color{black}87.96\%$\pm$1.85} & {\color{black}86.82\%$\pm$1.63} & {\color{black}82.21\%$\pm$1.06}   \\
$\tau$ = 0.1   & 88.78\%$\pm$1.96 & 86.36\%$\pm$1.39  &  83.04\%$\pm$1.02    \\
$\tau$ = 0.2   & 88.64\%$\pm$1.14 & 86.67\%$\pm$1.66  &  84.28\%$\pm$0.79    \\
$\tau$ = 0.5   & 90.07\%$\pm$1.81 & 86.98\%$\pm$1.73  &  84.89\%$\pm$0.95    \\
$\tau$ = 0.8   & {\color{black}87.85\%$\pm$2.13} & 85.58\%$\pm$1.90  &  83.70\%$\pm$0.91    \\
{\color{black}$\tau$ = 0.9} &  {\color{black}87.08\%$\pm$1.83} & {\color{black}83.26\%$\pm$2.03} & {\color{black}81.55\%$\pm$1.19} \\
{\color{black}$\tau$ = 0.99} & {\color{black}72.18\%$\pm$2.36} & {\color{black}73.41\%$\pm$2.69} & {\color{black}70.12\%$\pm$1.60} \\
\midrule
MS4: \\
$\tau \sim \mathcal{U}(0.1, 0.9)$   & 89.32\%$\pm$1.75 & 84.73\%$\pm$2.08  & 84.74\%$\pm$0.79 \\
$\tau \sim \mathcal{U}(0.3, 0.7)$   & \textbf{90.07}\%$\pm$1.33  & \textbf{87.37}\%$\pm$1.47 & 85.43\%$\pm$1.00 \\
\midrule
MS5: \\
$ \tau_{i} \sim \mathcal{U}(0.1, 0.3)$ \\
$ \Sigma_{i} \tau_{i} \sim \mathcal{U}(0.3, 0.7)$ & 89.53\%$\pm$0.87  & 86.13\%$\pm$2.15 &  \textbf{85.45}\%$\pm$0.62\\  
\bottomrule
\end{tabular}
\begin{tablenotes}
\footnotesize
\item $\mathcal{U}$ stands for uniform random number sampler.
\item $a,b$ are the two ends of a mask. $\tau$ is defined in Equation (\ref{Equ:mask_spectrum_ratio_tau}).
\end{tablenotes}
\end{threeparttable}
\label{Tab:AblaMaskingSchemes}
\end{table}

\subsubsection{Masking Schemes}\label{sec:masking_schemes}
{\color{black}
Following the protocol outlined in subsection~\ref{Subsec_evalprotocols}, we conducted a set of experiments to investigate different masking schemes discussed in Section~\ref{Sec:Subsection_masking_schemes}. The results are shown in Table~\ref{Tab:AblaMaskingSchemes}. For small masks ($\tau < 0.1$), the \textcolor{black}{ performance was} comparable to the baseline CNN\_full, which indicated that variations generated by masking-reconstruction were limited and contributed little to training. With large masks, ($\tau > 0.8$), the accuracy dropped significantly which indicated that the generated variations were inappropriate and contributed negatively to the training. 

Overall, the best masking scheme was ``MS4'' with $\tau \sim$ URN(0.3, 0.7) with randomized position and length of the mask. Note that MS1 performed significantly worse than the others. This agrees with the fact that some degree of randomization boosts the training of deep models.
}

\subsubsection{Joint vs. separate training of modules in TeaNet}
We investigated whether joint training of the reconstruction and classification modules in TeaNet is beneficial over a separate training. We ran the experiments for a varying amount of augmentation on all three databases.  The results in Table~\ref{Tab:AblaJointOrSep} show that joint training of the reconstruction and classification models significantly outperforms training them separately.

\begin{table}
\centering
\caption{Comparison of joint vs. separate training of two modules in TeaNet. $(\times n)$ corresponds to the number of masked samples generated for training the network}
\begin{tabular}{lcccc}
\toprule
\diagbox{Method}{Dataset}    & RRUFF\_IR & USGS & ReLab \\
\midrule
Separately($\times 5$)  & 86.94\%$\pm$2.06 & 84.81\%$\pm$1.56 & 83.57\%$\pm$1.21 \\
Jointly($\times 5$)     & 89.05\%$\pm$1.03 & 86.12\%$\pm$1.85 & 84.69\%$\pm$0.50\\
\midrule
Separately($\times 10$) & 87.96\%$\pm$0.92 &  84.65\%$\pm$1.38 & 83.05\%$\pm$1.22 \\
Jointly($\times 10$)    & 90.07\%$\pm$1.33 & 87.37\%$\pm$1.47  & 85.43\%$\pm$1.00  \\
\midrule
Separately($\times 20$) & 87.76\%$\pm$1.83 & 84.26\%$\pm$1.67 & 83.78\%$\pm$1.42\\
Jointly($\times 20$)    & 89.66\%$\pm$1.00 & 87.06\%$\pm$1.55 & 85.48\%$\pm$1.18\\
\bottomrule
\end{tabular}
\label{Tab:AblaJointOrSep}
\end{table}

{\color{black}
\subsubsection{Masking vs. Reconstruction}

In our approach, both components of masking and reconstruction are crucial and together they cooperate to generate artificial samples with sensible variations. Without masking, one would need to generate artificial samples from random noise which demands a large amount of data to train. As shown in Table~\ref{Tab:Main_Benchmark_Results}, GANs were even inferior to the baseline CNN\_Full. Without reconstruction, as showed in Table~\ref{Tab:Main_Benchmark_Results},  only masking e.g. CNN\_Partial or masking with rough reconstruction e.g. CutMix and MixUp, they all performed poorly. Masking followed by careful sensible reconstruction as in the proposed TeaNet indeed led to a significant improvement.
}

\addtolength{\tabcolsep}{-0.05cm}
\begin{table}[!h]
\centering
{\color{black}
\caption{Model complexity and the computational cost for the training of the proposed TeaNet compared to existing methods, excluding the classification model. For inference, only the classification model is to be used.}
\label{Tab_Complexity}
}
\begin{tabular}{lrcrr}
\toprule
{\color{black}Method}  & {\color{black}Model Size} & {\color{black}GPU Memo.} & {\color{black}Forw. Pass} & {\color{black}Back. Pass}\\
\midrule
{\color{black}LTSA~\cite{jenni2018self}} &  {\color{black}280.23 M} & {\color{black}5.12 G}  & {\color{black}112.17 ms}  & {\color{black}115.94 ms} \\
{\color{black}CWGAN~\cite{luo2018eeg}} & {\color{black}17.64 M}  & {\color{black}1.81 G} & {\color{black}4.39 ms}  & {\color{black}7.96 ms} \\
{\color{black}DAGAN~\cite{antoniou2017data}} & {\color{black}1.85 M} &  {\color{black}1.97 G} & {\color{black}33.95 ms} & {\color{black}101.24 ms} \\
{\color{black}TeaNet (ours)} & {\color{black}43.36 M} & {\color{black}2.06 G}  &  {\color{black}17.27 ms}  & {\color{black}40.64 ms} \\
\bottomrule
\end{tabular}
\end{table}
\addtolength{\tabcolsep}{0.05cm}

{\color{black}
\subsection{Model Complexity and Computational Cost}\label{sec_model_complexity}

In this section, we discuss the model complexity and the computational cost for the proposed methods with a comparison to existing generative model-based methods. As all the compared methods are proposed for data augmentation, and play no direct role in inference where only the classification is to be used, we calculated related metrics only for the training. Results can be found in Table~\ref{Tab_Complexity}. The required GPU memory and run-time were computed within an epoch with a batch size of 16 on a server with a NVIDIA GTX 3090 GPU and Intel(R) Xeon(R) CPU E5-2678.  
It can be seen that the proposed TeaNet used comparable GPU memory to DAGAN and CWGAN, ran much faster than DAGAN, \textcolor{black}{ and yet was slower than CWGAN.} \textcolor{black}{ We conclude that with the augmented samples, generated by reconstructing from randomly masked spectra, }we were able to train a relatively large model and achieved superior performance with computational costs that are comparable to the GANs.
}

\begin{figure*}[!htp]
	\centering
     \subfigure[Riebeckite vs Antigorite]{
	\includegraphics[width=0.45\textwidth]{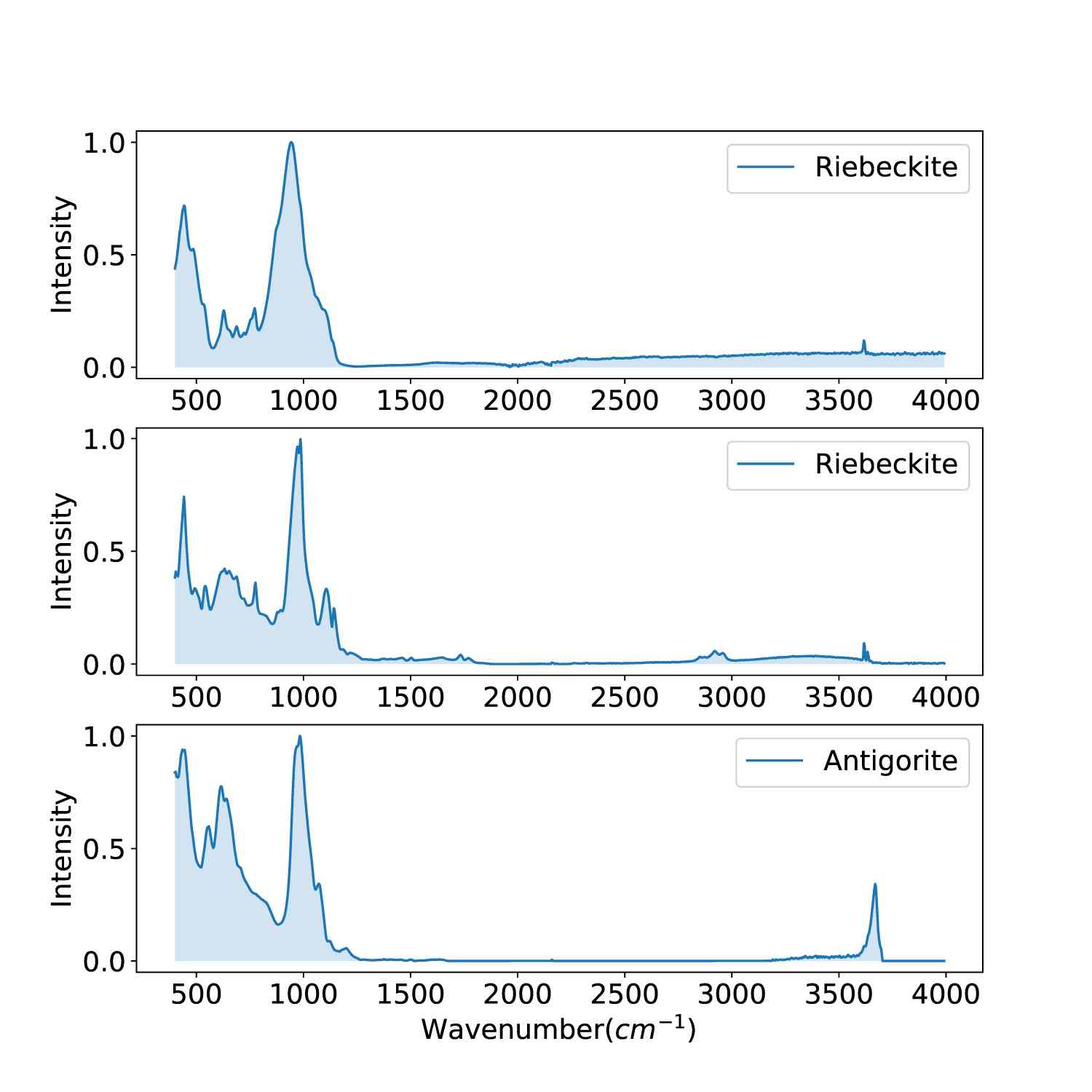}
    }
    \subfigure[Rhodonite vs Bustamite]{
	\includegraphics[width=0.45\textwidth]{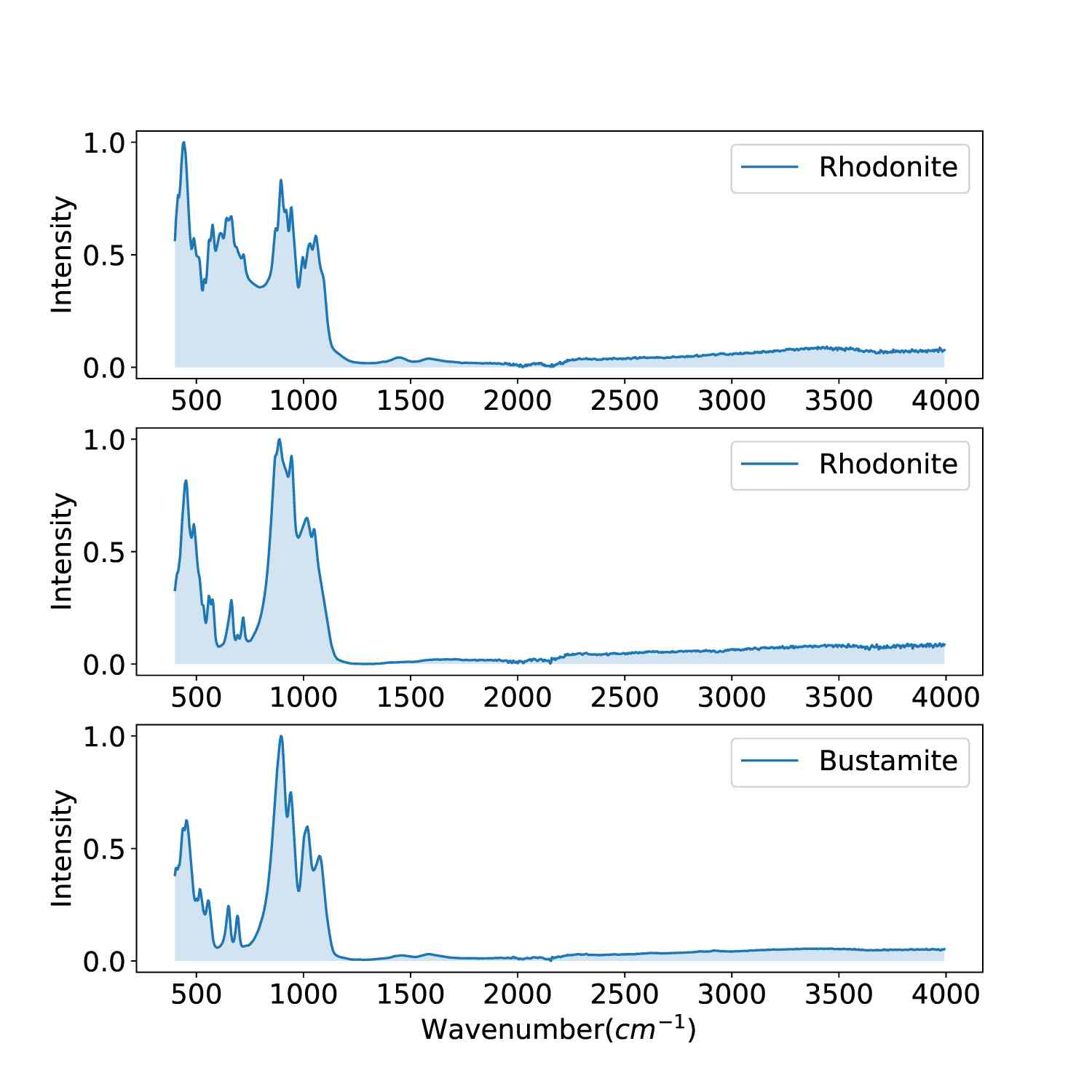}
    }
    \caption{Challenging cases where TeaNet succeeded while the state-of-the-art CNN failed. Each figure presents a case where the first and second rows plot samples in the training and test sets, respectively. The last row plots samples from the training set of the class misclassified by CNN. It can be seen that visually they are nearly indistinguishable.}
	\label{Fig:SuccessVsFailed_Examples}
\end{figure*}

\begin{table*}[!h]
\centering
\caption{Analysis of neuron responses of TeaNet and CNN on different regions of IR spectra of Riebeckite and Antigorite.}
\begin{threeparttable}
\begin{tabular}{lccccc}
\toprule
    & $R_{1}$ & $R_{2}$ & $R_{3}$ & $R_{4}$ & $R_{1-4}$ \\
\midrule
Wavenumber ($\text{cm}^{-1}$) & $400-475 $ & $476-600$ & $601-1200$ &$3600-3700$ & -\\
\midrule
Assigned Bands\cite{andrei2011_riebeckite} & \ce{T(Fe^{2+} - OH)} & \ce{bending Si-O-Si} & \ce{v_{s}(Si-O_{b}-Si)} & \ce{OH} stretching &  -  \\
& \ce{T(Mg^{2+} - OH)} & & \ce{v_{as}(Si-O_{b}-Si)} &   \\
&  & & \ce{v_{as}(O-Si-O)}  &   \\
\midrule
$\mathscr{S}$ of CNN  & 0.795 & 0.567 & 0.809 & 0.398 & {\color{black}0.748} $\downarrow$\\
\midrule
 $\mathscr{A}$ of CNN (\textit{Riebeckite})   & 0.066 & 0.054 & 0.446 & 0.010 & {\color{black}0.576} $\uparrow$ \\
\midrule
$\mathscr{A}$ of CNN (\textit{Antigorite})   & 0.082 & 0.094 & 0.355 & 0.055 & {\color{black}0.586} $\uparrow$\\
\midrule
$\mathscr{S}$ of TeaNet & 0.587 & 0.572 & 0.382 & 0.132 & \textbf{0.331} $\downarrow$\\
\midrule
$\mathscr{A}$ of TeaNet (\textit{Riebeckite}) & 0.278 & 0.117 & 0.505 & 0.019 & \textbf{0.919} $\uparrow$\\
\midrule
$\mathscr{A}$ of TeaNet (\textit{Antigorite}) & 0.086 & 0.130 & 0.611 & 0.136 & \textbf{0.963} $\uparrow$\\
\bottomrule
\end{tabular}
\begin{tablenotes}
\footnotesize
\item T - Transational modes; $v_{s}$ - Symmetric stretching; $v_{as}$ - Asymmetric stretching; $s_{nr}$ - Similarity of Neuron Responses; $\iota_{nr}$ - Intensity of Neuron Responses; $\downarrow$ - lower, better; $\uparrow$ - higher, better.
\end{tablenotes}
\end{threeparttable}
\label{Tab_RegionSimilarityAnalysis}
\end{table*}

\begin{table*}[!h]
\centering
\caption{Analysis of neuron responses of TeaNet and CNN on different regions of IR spectra of Rhodonite and Bustamite.}
\begin{threeparttable}
\begin{tabular}{lcccccc}
\toprule
    & $R_{1}$ & $R_{2}$ & $R_{3}$ & $R_{4}$ & $R_{5}$ & $R_{1-5}$ \\
\midrule
Wavenumber ($\text{cm}^{-1}$) & $400-550 $ & $551-750$ & $850-1100$ &$1600-1650$ & $3400-3450$ & -\\
\midrule
Assigned Bands\cite{RhodoniteAnalysis1966} & \ce{(M-O) stretching} & \ce{(Si-O-Si)} & \ce{(Si-O)} & \multicolumn{2}{c}{\ce{OH} bending and stretching} &  -  \\
& \ce{(Si-O) bending} & stretching & stretching &  && \\
\midrule
$\mathscr{S}$ of CNN  & 0.090 & 0.081 & 0.000 & 0.803 & 0.798 &  {\color{black}0.347} $\downarrow$ \\
\midrule
$\mathscr{A}$ of CNN (\textit{Rhodonite}) & 0.070 & 0.071 & 0.114 & 0.009 & 0.016 & {\color{black}0.281} $\uparrow$ \\
\midrule
$\mathscr{A}$ of CNN (\textit{Bustamite}) & 0.008 & 0.079 & 0.000 & 0.028 & 0.038 & {\color{black}0.153} $\uparrow$\\
\midrule
$\mathscr{S}$ of TeaNet & 0.325 & 0.054 & 0.533 & 0.000 & 0.000 & \textbf{0.392} $\downarrow$\\
\midrule
$\mathscr{A}$ of TeaNet (\textit{Rhodonite}) & 0.231 & 0.139 & 0.370 & 0.000 & 0.003 & \textbf{0.743} $\uparrow$ \\
\midrule
$\mathscr{A}$ of TeaNet (\textit{Bustamite}) & 0.136 & 0.145 & 0.710 & 0.000 & 0.000 & \textbf{0.992} $\uparrow$ \\
\bottomrule
\end{tabular}
\begin{tablenotes}
\footnotesize
\item $s_{nr}$ - Similarity of Neuron Responses; $\iota_{nr}$ - Intensity of Neuron Responses; $\downarrow$ - lower, better; $\uparrow$ - higher, better.
\end{tablenotes}
\end{threeparttable}
\label{Tab_RegionSimilarityAnalysis_Rhodonite}
\end{table*}

\subsection{Interpreting TeaNet on Challenging Cases}
Our experiments on real and synthetic data showed a clear advantage of TeaNet over other methods. \textcolor{black}{ We subsequently tried to interpret} how the TeaNet makes decisions in comparison to the baseline CNN. To this end, we focus on the most challenging cases of very similar samples that belong to different classes. Fig.~\ref{Fig:SuccessVsFailed_Examples} depicts two such examples that were correctly classified by TeaNet, but confused the CNN. Namely, the first and second rows in Fig.~\ref{Fig:SuccessVsFailed_Examples} show samples in the training and test sets respectively. The last row shows samples from the training set of the class misclassified by CNN. It can be seen that in both cases, the similarity between the spectra of the correct and the misclassified classes is very high, and therefore \textcolor{black}{it is} 
 challenging to distinguish between them. In fact, visually it is almost impossible for non-experts to make correct classification. 

By analysing these cases, we demonstrate a limited ability of CNN in identifying discriminant wavenumbers especially when dealing with very similar spectra. In contrast, TeaNet showed superior performance in locating discriminant regions and achieving high \textcolor{black}{classification accuracy} as we show below.

{\color{black}
To interpret how the networks make decisions, we visualize their neuron responses to spectra using Grad-CAM~\cite{Selvaraju2020}, and quantitatively evaluate what discriminant information \textcolor{black}{was} extracted and utilized for classification. To this end we define two measures; the \textit{similarity of two responses} $\mathscr{S}$ and the \textit{attention on discriminant wavenumbers} $\mathscr{A}$. 
}
\begin{align}
    \mathscr{S} = \frac{\textbf{v}_{r} \cdot \textbf{v}_{s}}{ \| \textbf{v}_{r} \| \| \textbf{v}_{s} \|}
\end{align}
where $\textbf{v}_{r}$ and $\textbf{v}_{s}$ are two responses/vectors which correspond to a segment or a full spectrum. $\mathscr{S}$ measures how similar two responses are. For instance, $\mathscr{S} = 1$ means the corresponding network reacts almost constantly to any spectra, which of course may lead to a very low classification accuracy. 

\begin{align}
    \mathscr{A}_{j} = \frac{\| \textbf{v}_{j} \|_{1} }{\| \textbf{v} \|_{1} }  
\end{align}
where $\textbf{v}_{j}$ is the $j^{th}$ segment of a spectrum $\textbf{v}$, and  $\mathscr{A}_{j} \in (0, 1]$ is the normalized sum of the intensity of a response on the $j^{th}$ segment (region). $\mathscr{A}_{j}$ on discriminant regions/segments indicates how the intensity/energy is distributed over the spectrum. Ideally, 
{\color{black} 
high values of $\mathscr{A}$ should correspond to the discriminative segments in the input, which in turn should correlate with improved classification accuracy.} 

\begin{figure*}
	\centering{
	\includegraphics[width=0.9\textwidth]{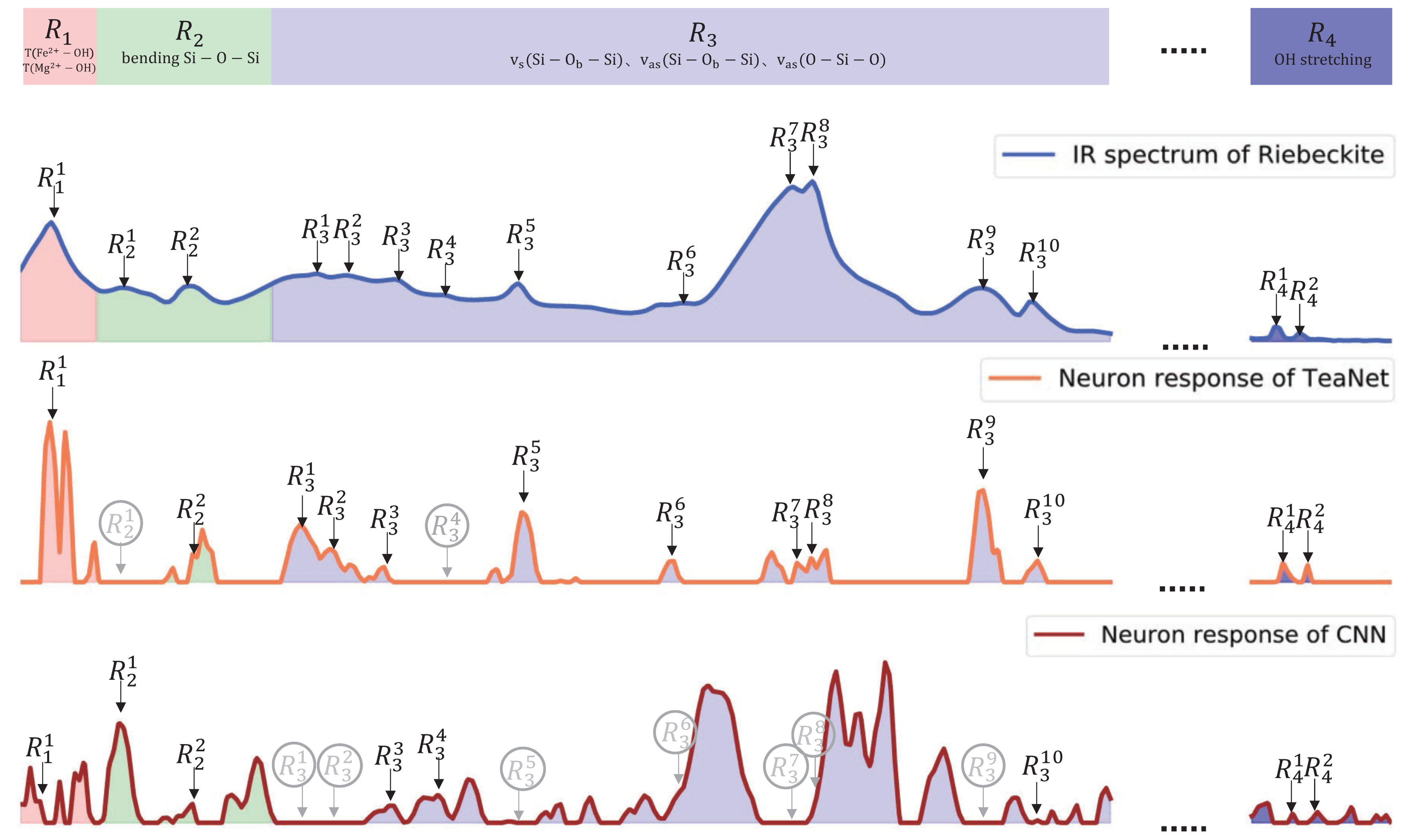}
    }
    \caption{Comparison of neuron responses of TeaNet and CNN on a challenging case of discriminating between ``Riebeckite'' and ``Antigorite'', the spectra of which, see Fig.~\ref{Fig:SuccessVsFailed_Examples}, is almost indistinguishable visually from one another. On this case, TeaNet succeeded while the CNN failed. According to \cite{andrei2011_riebeckite}, we categorized the peaks into $\{R_{i}, i=1,\cdots, 4\}$ and marked the peaks as $R_{i}^{k}$ where $k$ is the index of the peak within its region for convenience. We visualize the neuron responses(activation maps) of TeaNet and CNN by Grad-CAM, and highlighted the matches to the peaks of the spectrum. The peaks which were ignored by either method were circled and marked in grey. It can be seen that responses of TeaNet to almost all of the characteristic peaks are accurate and well-shaped, while CNN missed quite a few.}
	\label{Fig:ChallengingCase_Gradcam}
\end{figure*}

\begin{figure*}
	\centering{
	\includegraphics[width=0.9\textwidth]{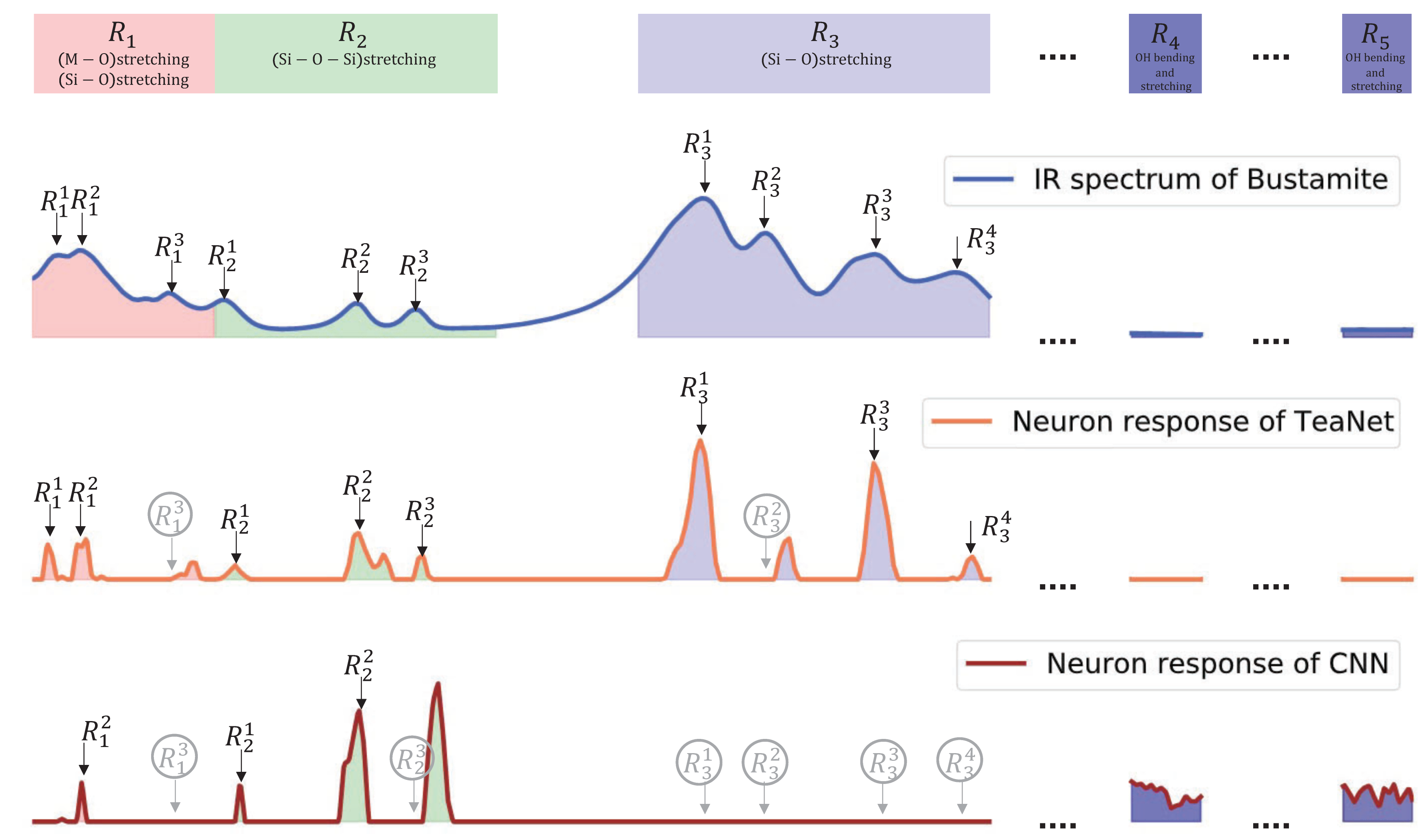}
    }
    \caption{Comparison of neuron responses of TeaNet and CNN on a challenging case of discriminating between ``Riebeckite'' and ``Antigorite'', the spectra of which, see Fig.~\ref{Fig:SuccessVsFailed_Examples}, is almost indistinguishable visually from one another. On this case, TeaNet succeeded while the state-of-the-art CNN failed. According to \cite{RhodoniteAnalysis1966}, we categorized the peaks into $\{R_{i}, i=1,\cdots, 5\}$ and marked the peaks as $R_{i}^{k}$ where $k$ is the index of the peak within its region for convenience. We visualize the neuron responses(activation maps) of TeaNet and CNN by Grad-CAM, and highlighted the matches to the peaks of the spectrum. The peaks which were ignored by either method were circled and marked in grey. It can be seen that responses of TeaNet to almost all of the characteristic peaks are accurate and well-shaped, while CNN missed quite a few. 
    }
	\label{Fig:ChallengingCase_Gradcam_2}
\end{figure*}

{\color{black}
Due to space limitations, we only present detailed analysis for two challenging examples shown in Fig.~\ref{Fig:SuccessVsFailed_Examples}. Additional examples of challenging cases are shown in supplementary information. 

\subsubsection{\textbf{Antigorite vs Riebeckite}}

Fig.~\ref{Fig:SuccessVsFailed_Examples}(a) shows the spectra of the minerals \textit{Riebeckite} and \textit{Antigorite},
the chemical formula of which are 
\ce{Na_{2}(Fe^{2+}_{3}Fe^{3+}_{2})Si_{8}O_{22}(OH)_{2}}
and 
\ce{Mg_{3}Si_{2}O_{5}(OH)_{4}} respectively. 
According to \cite{andrei2011_riebeckite}, we identified all the discriminant regions (wavenumbers) on the spectra of Riebeckite. We categorized all the characteristic peaks into four regions $\{R_{i}, i=1,\cdots, 4\}$ and marked the peaks as $R_{i}^{k}$ where $k$ is the index of the peak within its region.

Fig. \ref{Fig:ChallengingCase_Gradcam} shows the discriminant peaks on the spectrum and the corresponding matches (peaks) on the neuron responses of TeaNet and CNN, which were found as the closest peak within a small neighborhood, $0.5\%$ of the spectrum length, of the discriminant peaks on the original spectrum. Peaks which were ignored by either method were circled and marked in grey. It can be seen that responses of TeaNet
to almost all of the characteristic peaks are accurate and well-shaped, while CNN missed 7 out of 15 peaks. The ability to capture more discriminant peaks enables TeaNet to distinguish spectra with high similarity. 

Further more, to quantitatively analyse the neuron responses, we calculated $\mathscr{S}$ and $\mathscr{A}$ on each region of $\{R_{i}, i=1,\cdots 4 \}$ individually as well as the combination of these regions. Results are shown in Table \ref{Tab_RegionSimilarityAnalysis}. It can be seen that $\mathscr{S}$ of TeaNet on full spectra was 0.331 which is significantly lower than 0.748 of CNN. This indicates that compared to CNN, TeaNet reacted differently to spectra of different minerals which enables it to distinguish spectra with minor differences. $\mathscr{A}$ of CNN on the regions combined was less than $60\%$, while that of TeaNet was more than $90\%$.  This demonstrated that compared to CNN, TeaNet could better identify the discriminant regions, make predictions based upon those selected wavenumbers and therefore achieve higher accuracy.

\subsubsection{\textbf{Bustamite vs Rhodonite}}
Fig. \ref{Fig:SuccessVsFailed_Examples}(b) shows the spectra of the minerals, \textit{Bustamit} and \textit{Rhodonite}, the chemical formula of which are \ce{Mn_{2}Ca_{2}MnCa(Si_{3}O_{9})_{2}} and \ce{CaMn_{3}Mn(Si_{5}O_{15})} respectively. 

Similar to Fig.~\ref{Fig:ChallengingCase_Gradcam}, we plot discriminant peaks on the original spectrum and their matches on neuron responses in Fig.~\ref{Fig:ChallengingCase_Gradcam_2}. Evidently TeaNet captured much more discriminant peaks than CNN. 

We calculated $\mathscr{S}$ and $\mathscr{A}$ on full spectra as well as 5 discriminant regions~\cite{RhodoniteAnalysis1966}. Results are presented in Table \ref{Tab_RegionSimilarityAnalysis_Rhodonite}. It can be seen that in this case $\mathscr{S}$ of TeaNet and CNN are 0.392 and 0.347 respectively, which indicates both reacted differently to the compared minerals. On the other hand, $\mathscr{A}$ of CNN to both minerals are 28.1\% and 15.3\% respectively, which is extremely low and indicates that CNN actually failed to concentrate on the discriminant regions. Indeed, by checking the neuron responses, we observed that when fed with Bustamite, CNN completely missed the discriminant \textcolor{black}{region} $R_{3}$. 

}

This confirms that CNN's ability to locate the discriminant regions is limited, especially for challenging cases as the above-mentioned ones . While enhanced by the task of reconstruction of masked spectra, even with the same network architecture, TeaNet showed great potential in identifying discriminant wavenumbers and achieved superior performance of classification. In this sense, TeaNet seems to be able to implicitly perform wavenumbers selection, a technique which has been widely used in industrial applications of vibrational spectroscopy.

\section{Conclusion and Future Work}\label{sec:conclusion}
In this paper, we propose a task-enhanced augmentation network for a few-shot vibrational spectrum recognition. By reconstructing randomly masked samples, the network is encouraged to generate augmented samples with meaningful variation in the masked regions. Results on both synthetic and real-world datasets verified the superiority of the proposed method. Moreover, by visualizing and analysing the neuron responses of TeaNet, we demonstrated TeaNet's excellent ability of identifying discriminant wavenumbers, which allows it to distinguish spectra with high similarity in challenging cases where the state-of-the-art methods including CNN tend to fail.  Future research will investigate additional losses which measure the diversity of artificial samples and learnable masking parameters to further improve the diversity and efficiency of the sample-generating process, and ultimately increase the classification accuracy.

}

\section*{Acknowledgment}
This research was supported by National Natural Science Foundation of China No. 62076140.

\bibliographystyle{IEEEtran}
\bibliography{cai_fewshot}

\end{document}